\documentclass[letterpaper]{article} 
\usepackage{aaai2026}  
\usepackage{times}  
\usepackage{helvet}  
\usepackage{courier}  
\usepackage[hyphens]{url}  
\usepackage{graphicx} 
\usepackage{natbib}  
\usepackage{caption} 
\usepackage{algorithm}
\usepackage{algorithmic}

\usepackage{newfloat}
\usepackage{listings}
\DeclareCaptionStyle{ruled}{labelfont=normalfont,labelsep=colon,strut=off} 
\floatstyle{ruled}
\newfloat{listing}{tb}{lst}{}
\floatname{listing}{Listing}
\usepackage{amsmath}
\usepackage{amssymb}
\usepackage{tabularx}
\usepackage{subcaption}
\usepackage{adjustbox}
\usepackage{svg}
\usepackage{stmaryrd}
\usepackage{booktabs}
\usepackage{caption}
\usepackage{multirow}
\usepackage{svg}
\usepackage{pifont}
\usepackage{fancybox}
\usepackage{makecell}
\usepackage{colortbl}
\usepackage{xcolor}
\usepackage{amsfonts}
\usepackage[capitalize]{cleveref}
\definecolor{mygray}{gray}{.95}
\usepackage{soul}

\title{Transferable Model-agnostic Vision-Language Model Adaptation \\ for Efficient Weak-to-Strong Generalization}
\author {
    Jihwan Park\textsuperscript{\rm 1},
    Taehoon Song\textsuperscript{\rm 2},
    Sanghyeok Lee\textsuperscript{\rm 2},
    Miso Choi\textsuperscript{\rm 1},
    Hyunwoo J. Kim\textsuperscript{\rm 2}\thanks{Corresponding author.}
}
\affiliations{
    \textsuperscript{\rm 1}Korea University \hspace{0.4cm} \\
    \textsuperscript{\rm 2}KAIST \\

    \{jseven7071,  
    miso8070\}@korea.ac.kr \\
    \{taehoons,
    sanghyeoklee,
    hyunwoojkim\}@kaist.ac.kr
}

\DeclareMathOperator*{\argmax}{arg\,max}
\DeclareMathOperator*{\argmin}{arg\,min}
\newcommand{\R}{\mathbb{R}}

\newcommand{\xmark}{\text{\ding{55}}}
\newcommand{\taskclass}{C_\text{task}}

\newcommand{\ie}{\textit{i.e., }}
\newcommand{\eg}{\textit{e.g., }}

\begin{document}

 \maketitle


\begin{abstract}
Vision-Language Models (VLMs) have been widely used in various visual recognition tasks due to their remarkable generalization capabilities. 
As these models grow in size and complexity, fine-tuning becomes costly, emphasizing the need to reuse \textbf{adaptation knowledge} from `weaker' models to efficiently enhance `stronger' ones.
However, existing adaptation transfer methods exhibit limited transferability across models due to their model-specific design and high computational demands.
To tackle this, we propose \textbf{Trans}ferable \textbf{M}odel-agnost\textbf{i}c adap\textbf{ter} (\textbf{TransMiter}), a light-weight adapter that improves vision-language models `without backpropagation'.
TransMiter captures the knowledge gap between pre-trained and fine-tuned VLMs, in an `unsupervised' manner.
Once trained, this knowledge can be seamlessly transferred across different models without the need for backpropagation.
Moreover, TransMiter consists of only a few layers, inducing a negligible additional inference cost.
Notably, supplementing the process with a few labeled data further yields additional performance gain, often surpassing a fine-tuned stronger model, with a marginal training cost.
Experimental results and analyses demonstrate that TransMiter effectively and efficiently transfers adaptation knowledge while preserving generalization abilities across VLMs of different sizes and architectures in visual recognition tasks.
\end{abstract}


\section{Introduction}
The rapid evolution of vision-language models (VLMs)~\cite{align,clip} has driven significant advancements in computer vision. 
These models are typically built with two modality-specific encoders, one for image and one for text, and are trained on large datasets of image-text pairs using contrastive loss. 
Here, modality-specific encoders learn to map each modality input into a joint embedding space, enabling zero-shot capabilities across a wide range of visual recognition tasks.
The broad and generalized knowledge embedded in VLMs enables efficient adaptation on downstream tasks with few labeled data, as explored in~\cite{lweib,hpt,zhou2022coop,zhou2022cocoop,promptsrc,khattakMaPLe,li2024promptkd}.
However, as VLMs are scaled~\cite{internvl,openclip} or equipped with more complex architectures~\cite{coca}, fine-tuning them requires massive memory and computational costs.

\begin{figure}[tp]
\includegraphics[width=\linewidth]{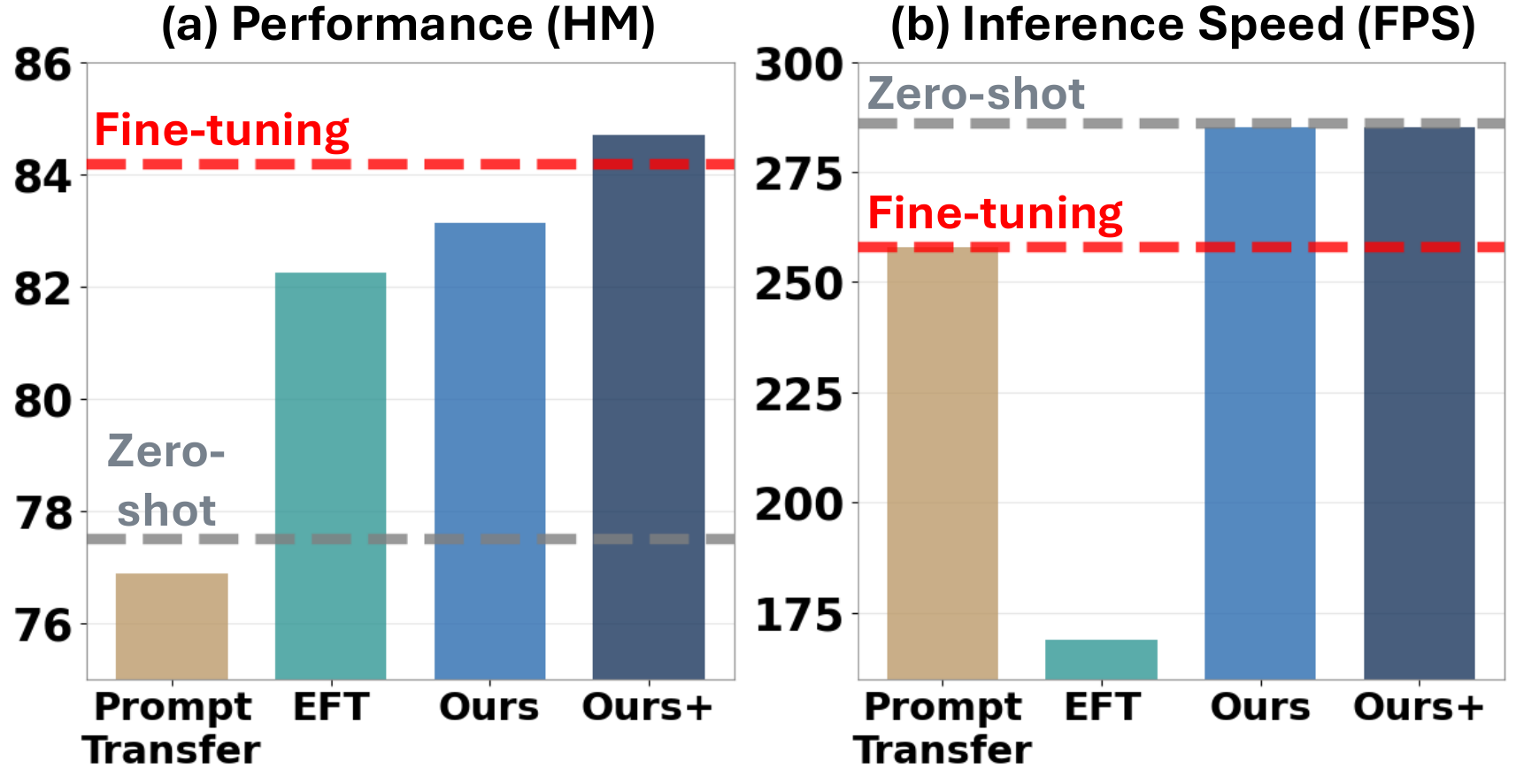}
{\caption{\textbf{Comparison of adaptation knowledge transfer methods.}
Performance is averaged over 11 visual recognition tasks in a base-to-novel setting.
\textbf{TransMiter} (ours) outperforms other adaptation transfer approaches, including Prompt Transfer~\cite{prompt_trans} and EFT~\cite{Emulator}, while maintaining inference speed nearly identical to a {zero-shot} model (gray), which serves as the upper-bound.
With a small amount of labeled data, \textbf{TransMiter+} (ours+) surpasses its supervised counterpart~\cite{promptsrc}, \eg Fine-tuning (red).}}
\label{fig:comparison_fig1}
\end{figure}

In this context, an important research question arises; ``How can the knowledge gained from adapting smaller, weaker models, \ie{\em adaptation knowledge}, be effectively leveraged to improve larger and more advanced models?'' 
One common approach~\cite{prompt_trans,steerLLM,proxy_tuning,Emulator} involves directly reusing a subset of parameters from a weaker model to a stronger one.
However, due to semantic discrepancy or dimensional mismatch between different models, additional training is often necessary~\cite{prompt_trans,steerLLM} to align their internal representations properly.
Some works~\cite{lu2023inference,proxy_tuning,Emulator,ensemblevlm} mitigate this issue by enforcing semantic consistency (\eg model prediction), reducing the need for re-training but at the cost of slower inference due to multiple model usage.
Alternatively, knowledge distillation~\cite{yuan2020revisiting,reverse_kd,weak_to_strong,li2024promptkd} has been demonstrated as an effective way for smaller models to guide larger ones.
Still, this approach requires expensive retraining whenever a new model is introduced, limiting its practicality for rapidly evolving models.

In this work, we aim to design an adaptation module to efficiently transfer \textbf{adaptation knowledge} from the weaker to stronger VLMs.
Our design is guided by three key objectives: (1) \textbf{model-agnostic compatibility}, (2) \textbf{computationally efficient transfer}, and (3) \textbf{minimal additional inference cost}.
To this end, we propose \textbf{Trans}ferable \textbf{M}odel-agnost\textbf{i}c adap\textbf{ter} (\textbf{TransMiter}), a light-weight adapter that efficiently transfers the adaptation knowledge across different models `without backpropagation'. 
Instead of directly transferring knowledge, we adopt a novel learning paradigm \textbf{“knowledge extraction and transfer”}.
TransMiter first extracts {adaptation knowledge} by learning to capture the knowledge gap through the differences in predicted logits between pre-trained and fine-tuned weak models in an unsupervised manner.
Once extracted, TransMiter enables seamless transfer to any stronger models by leveraging the consistent dimensionality and semantics of logits.

To effectively harness the adaptation knowledge, we utilize auxiliary classes alongside task classes to increase the dimensionality of logit vectors. 
It enhances the expressive power of the logits, thereby boosting the acquisition of adaptation knowledge and transferability across models.
Additionally, we propose a basis change to further align models within the latent space.
The basis for the transferred model is derived through a closed-form solution, thus eliminating the need for computation-intensive backpropagation.
By incorporating the above two methodologies, TransMiter achieves exceptional transferability.
After the adaptation knowledge transfer process, TransMiter itself becomes a strong initialization point, enabling a small amount of labeled data to deliver substantial performance boosts, often surpassing fine-tuned stronger counterparts, while incurring only minimal additional training cost.
Notably, its light-weight design, consisting of simple projection matrices and a single MLP layer, ensures minimal additional inference cost, making it both efficient and scalable.

Extensive experiments demonstrate that TransMiter effectively transfers adaptation knowledge across models of varying sizes and architectures, confirming its superior transferability in diverse visual recognition tasks.
The contributions of TransMiter can be summarized as:
\begin{itemize}
    \item We propose \textbf{Trans}ferable \textbf{M}odel-agnost\textbf{i}c adap\textbf{ter} (\textbf{TransMiter}), a light-weight adapter for enhancing vision-language models without backpropagation.
    \item We incorporate two key techniques, auxiliary class expansion and basis change, that significantly boost the transferability of TransMiter, while the addition of a few labeled data further unlocks its full potential.
    \item We demonstrate TransMiter's effectiveness and efficiency in transferring adaptation knowledge across a wide range of datasets and models.
\end{itemize}

\section{Related Works}
\textbf{Vision-Language Model Adaptation.} Vision-language models (VLMs), such as CLIP~\cite{clip}, have demonstrated strong generalization abilities across various visual recognition tasks.
Leveraging this strength, numerous studies have investigated fine-tuning VLMs on visual recognition tasks, such as prompt learning~\cite{lweib,hpt,zhou2022coop,zhou2022cocoop,prograd,rpo,dapt,promptsrc,khattakMaPLe,li2024promptkd,vdrp}, low-rank adaptation~\cite{lora_vlm}, linear probing~\cite{lfa}, and regularization~\cite{prometar,prograd,promptsrc}.
However, existing VLM adaptation methods often neglect the models' rapid evolution in terms of size and complexity, resulting in the inconvenience of retraining~\cite{prompt_trans} whenever a new model comes.
To address this limitation, we focus on developing a transferable adaptation module for VLMs, enabling the efficient transfer of adaptation knowledge acquired during fine-tuning across different VLMs, regardless of their size or architectural design.

\noindent \textbf{Weak-to-Strong Generalization.}
In an era of rapidly evolving VLMs, {``weak-to-strong generalization''}, where stronger models leverage the knowledge of weaker models, is emerging as a new research challenge for enabling efficient model adaptation.
Previous methods try to transfer either a portion of a model’s internal parameters~\citep{prompt_trans,TLT,steerLLM} or the entire model itself~\citep{lu2023inference,proxy_tuning, Emulator} to other models.
However, differences in architecture often hinder parameter reuse, and even when feasible, aligning representation spaces requires costly retraining.
While full-model transfer alleviates the problem using predicted logits, it requires running multiple models, leading to significant inference cost.
Some other studies~\cite{yuan2020revisiting,reverse_kd,weak_to_strong,li2024promptkd} have shown that weaker models can effectively supervise stronger ones, a concept known as reverse knowledge distillation~\cite{reverse_kd}. 
While these methods incur no additional inference cost, they still require expensive retraining whenever a new model is introduced, making it impractical for frequent model updates.
In our work, we design an adaptation module for high cross-model transferability and fast inference speed.
Additionally, we employ ``knowledge extraction and transfer'' approach, where adaptation knowledge is extracted only ``once'' from a weaker model and efficiently transferred to a stronger model.

\section{Preliminaries}
\label{subsec:3.1}
\textbf{Vision-Language Model.}  A pre-trained Vision-Language Model (VLM), denoted as $\theta_\text{pt}=\{\mathcal{V}_\text{pt},\mathcal{T}_\text{pt}\}$, is trained on large-scale image-text datasets using a contrastive loss, where $\mathcal{V}_\text{pt}$ and $\mathcal{T}_\text{pt}$ denote the pre-trained image encoder and text encoder, respectively.
Given an input image $x$ and task classes $C_\text{task}$ are processed by the image and text encoders, producing an image feature and a set of text features, one for each class $c\in C_\text{task}$.
These text features serve as ``anchors'', acting as reference points for classification.
To determine the predicted class $\hat{y}$, the image feature is compared to each class's text feature using a similarity measure. 
The class with the highest similarity to the image feature is selected as the predicted class. 
This process is expressed as:
\begin{equation}
    \begin{split}
        z_\text{pt}(c;x)&=\textbf{sim}\left(\mathcal{V}_\text{pt}(x),\mathcal{T}_\text{pt}(c)\right),\forall c\in\taskclass\\
        \hat{y}&={\argmax}_{c\in\taskclass}z_\text{pt}(c;x),
    \end{split}
    \label{eq:pred}
\end{equation}
where $\textbf{sim}$ denotes the cosine similarity. 
We will omit the $(c;x)$ terms in logits for brevity.

 To incorporate task knowledge into VLMs, the model can be fine-tuned using contrastive loss between image and text features, yielding a fine-tuned model $\theta_\text{ft} = \{\mathcal{V}_\text{ft}, \mathcal{T}_\text{ft}\}$.

\noindent \textbf{Problem Setting.}
Suppose that we have a pre-trained source model $\theta_\text{pt-s}$ (\eg weak model) and its fine-tuned counterpart $\theta_\text{ft-s}$.
Our goal is to capture the knowledge gap, \ie adaptation knowledge, between the pre-trained and fine-tuned source models, then transfer this knowledge to the pre-trained target model $\theta_\text{pt-t}$ (\eg strong model) without using any labeled data.
Specifically, the extraction of adaptation knowledge $\delta_\text{s}$ can be expressed as:
\begin{equation}
    \begin{split}
        \delta_\text{s} = \mathcal{E}\left(\theta_\text{ft-s}, \theta_\text{pt-s}\right),\\
    \end{split}
    \label{eq:adp_extract}
\end{equation}
where $\mathcal{E}$ is the adaptation knowledge extraction operation.

Once the adaptation knowledge is extracted, it is transferred to the pre-trained target model, resulting in a knowledge-enhanced target model $\theta_\text{t}^*$ as:
\begin{equation}
    \begin{split}
        \theta_\text{t}^*=\mathcal{J}\left(\theta_\text{pt-t}, \delta_\text{s}\right),\\
    \end{split}
    \label{eq:adp_transfer}
\end{equation}
where $\mathcal{J}$ is the adaptation knowledge transfer operation.

 The enhanced target model $\theta_\text{t}^*$ is expected to obtain the effects of fine-tuning in two perspectives: (1) preservation of its pre-trained knowledge, and (2) superiority over the fine-tuned source model for its usefulness.
We can formalize our objective as follows:
\begin{equation}
    \begin{split}
        \max\left(\mathcal{P}(\theta_{\text{pt-t}}), \mathcal{P}(\theta_{\text{ft-s}})\right)  \leqslant \mathcal{P}(\theta_\text{t}^*)
        \lesssim \mathcal{P}(\theta_\text{ft-t}),
    \end{split}
    \label{eq:adp_eval}
\end{equation}
where $\mathcal{P}(\theta)$ refers to evaluation performance of model $\theta$.
The target model fine-tuned with labeled data, $\mathcal{P}(\theta_\text{ft-t})$, serves as a theoretical upper bound~\cite{weak_to_strong}.

\section{Method}
We introduce the \textbf{TRANS}ferable \textbf{M}odel-agnost\textbf{I}c adap\textbf{TER} (\textbf{TransMiter}), a light-weight adapter which efficiently enhances stronger model by leveraging weaker model's adaptation knowledge without backpropagation. 
We provide TransMiter's architectural design and the adaptation knowledge extraction process $\mathcal{E}$ in \cref{subsec:3.2}.
We then present its forward-only adaptation knowledge transfer process $\mathcal{J}$, which facilitates efficient adaptation transfer, in \cref{sec:3.3}.
\subsection{Transferable Model-Agnostic Adapter}
\label{subsec:3.2}

\begin{figure*}[t!]
    \centering
    \includegraphics[width=0.9\textwidth]{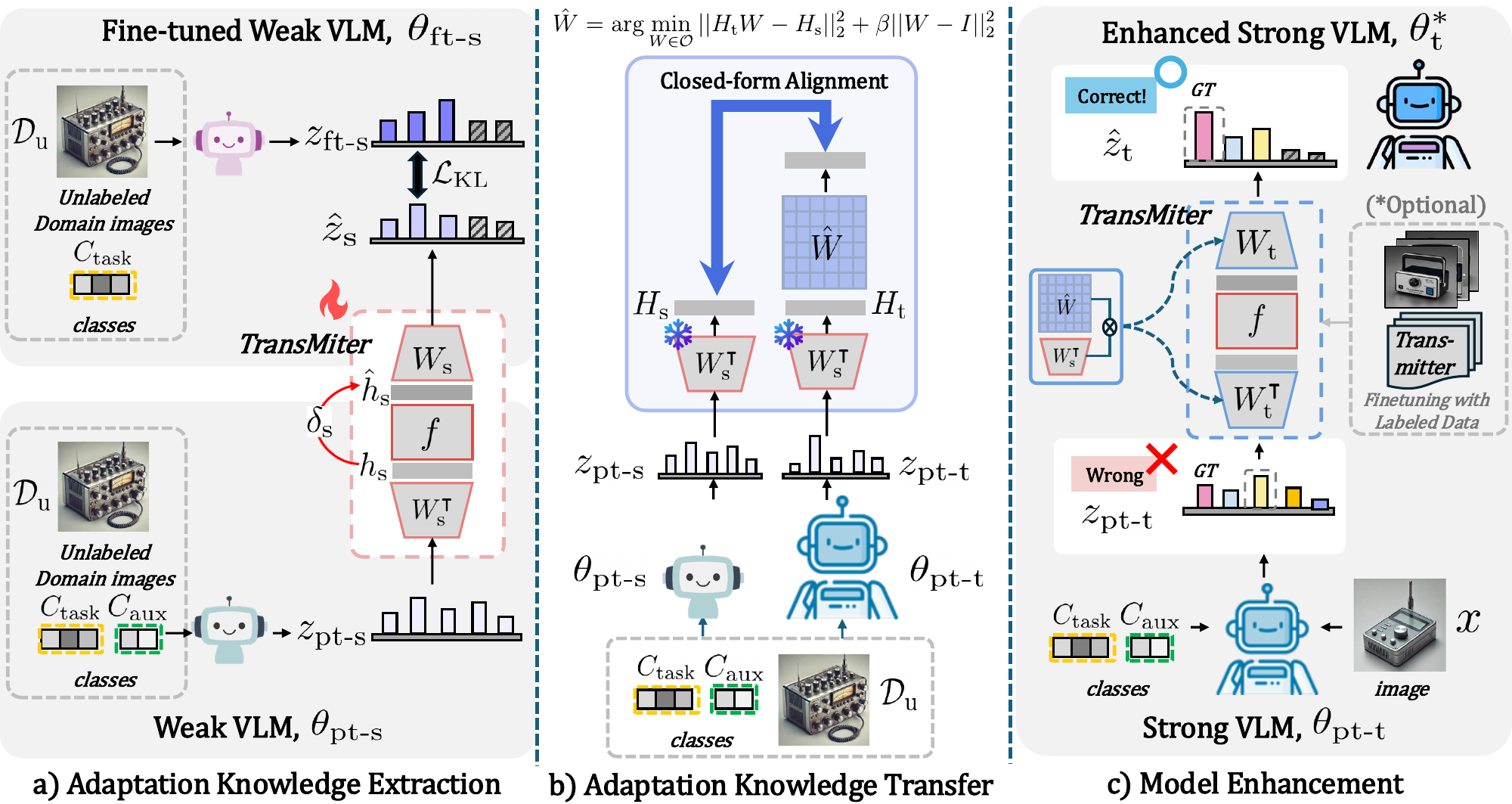}
    \caption{\textbf{Overall pipeline.}
    \textbf{(a) Adaptation Knowledge Extraction.} Given pre-trained $\theta_\text{pt-s}$ and fine-tuned weak VLMs $\theta_\text{ft-s}$, TransMiter captures the adaptation knowledge $\delta_\text{s}$, by minimizing the distance between the refined logits $\hat{z}_\text{s}$ and the fine-tuned weak VLM logits $z_\text{ft-s}$.
    The adapter takes the zero-shot model logits as input and incorporates both task classes $C_\text{task}$ and auxiliary classes $C_\text{aux}$.
    \textbf{(b) Adaptation Knowledge Transfer.} Once the strong VLM $\theta_\text{pt-t}$ is available, the mapping matrix $\hat{W}$ is computed using a closed-form solution to align the input features between the weak ($H_\text{s}$) and strong ($H_\text{t}$) VLMs, replacing the original transition matrix $W_\text{s}$ with $W_\text{t}=\hat{W}^\intercal W_\text{s}$.
    \textbf{(c)} \textbf{Model Enhancement.} During the inference with the target VLM using TransMiter $\theta_\text{t}^*$, the pre-trained target VLM logits $z_\text{pt-t}$ are passed through the adapter, resulting in enhanced predictions.
    Subsequently, as TransMiter offers a strong initial point, it can be fine-tuned with labeled data to maximize its capability.
    }
\label{fig:method} 
\end{figure*}

Previous adaptation techniques~\cite{zhou2022coop,khattakMaPLe,zhou2022cocoop, prograd, promptsrc, li2024promptkd, lfa} rely on internal model parameters that differ across VLMs, making direct transfer across different VLMs challenging.
To mitigate this problem, our adapter uses the predicted logits from the VLM as input.
The logits exhibit two key properties: (1) a fixed dimensionality that aligns with the number of classes, and (2) semantic consistency for each logit element. 
These properties facilitate effective communication~\cite{relrep} between different models, enabling seamless reuse of our adapter.
Moreover, our adapter does not require access to the parameters of VLM, thereby achieving high efficiency.

\noindent  \textbf{Prediction-based Adapter.}
Given an image $x$, we start with the input logits $z_\text{pt-s} \in \mathbb{R}^{N_\text{task}}$ obtained from the pre-trained weak VLM $\theta_\text{pt-s}$, where $N_\text{task}$ refers to the number of task classes.
These logits are projected into a $D$-dimensional latent space by multiplying them with the projection matrix $W_\text{in}\in\R^{N_\text{task}\times D}$, producing the input feature $h_\text{s}$. 
Then the input feature $h_\text{s}$ is passed through a transformation function $f:\R^{D}\rightarrow\R^{D}$, which consists of a single MLP layer with a residual connection, yielding $\hat{h}_\text{s}$.
 Finally, the output feature $\hat{h}_\text{s}$ is projected back into the original logit space by multiplying them with a reconstruction matrix $W_\text{out}\in\R^{D\times N_\text{task}}$.
As a result, we can obtain refined logits of the weak VLM $\hat{z}_\text{s}$, which can be written as:
\begin{equation}
    \begin{split}
        z_\text{pt-s}&=[\textbf{sim}(\mathcal{V}_\text{pt-s}(x),\mathcal{T}_\text{pt-s}(c))]_{c\in C_\text{task}}\in\R^{N_\text{task}},\\
         h_\text{s} &= z_\text{pt-s}W_\text{in}\in\R^{D}, \\
         \hat{h}_\text{s}&=f(h_\text{s})\in\R^{D},\\
         \hat{z}_\text{s}&=\hat{h}_\text{s}W_\text{out}\in\R^{N_\text{task}}, \\
    \end{split}
\label{eq:upa_pre}
\end{equation}
where $f(h)=h+\text{MLP}(h)$.

 A key consideration is adapting to the task while preserving the generalized knowledge in the original model. 
Since TransMiter employs simple 2D matrices to facilitate transitions between the logit and latent space, we enforce an inverse relationship between the projection and reconstruction matrices. 
If the transformation function $f$ operates as an identity function (\eg, $f(h)=h$), the output remains identical to the original logits, making an inverse relationship between the matrices reasonable. 
Here, we adopt orthogonality for the two matrices, \ie $W_\text{in}W_\text{in}^\intercal=W_\text{out}^\intercal W_\text{out}=I$, and set the two matrices in a transposed relationship, \ie $W_\text{in}=W_\text{out}^\intercal$.
These properties provide the projection and reconstruction matrices with an implicit regularization effect~\cite{ortho_OP,gain_orth,OFT} and establish an inverse relationship, \ie $W_\text{in}W_\text{out}=I$, between them.
For simplicity, we will denote $W_\text{out}$ as $W_\text{s}$, representing the transition matrix for the weak VLM.
We can rewrite \eqref{eq:upa_pre} in a simplified form as follows:

\begin{equation}
    \begin{split}
        \hat{z}_\text{s} = f(z_\text{pt-s}W_\text{s}^\intercal)W_\text{s}\in\R^{N_\text{task}}.\\
    \end{split}    
    \label{eq:upa}
\end{equation}

 The training objective for TransMiter is to ensure that our refined prediction $\hat{z}_\text{s}$ closely mimics the prediction of the fine-tuned weak VLM, $z_\text{ft-s}$. 
By doing so, we can obtain the adaptation knowledge $\delta_\text{s}$ from the fine-tuned model in an unsupervised manner, leveraging unlabeled data $\mathcal{D}_\text{u}$.
With the KL divergence loss, the training objective can be written as:
\begin{equation}
\label{eq:loss}
    \begin{split}
        z_\text{ft-s}&=[\textbf{sim}(\mathcal{V}_\text{ft-s}(x),\mathcal{T}_\text{ft-s}(c))]_{c\in C_\text{task}}\in\R^{N_\text{task}},\\
         \mathcal{L}&=\text{KL}\left(\sigma(z_\text{ft-s}/\tau_\text{ft-s}),\sigma(\hat{z}_\text{s}/\tau_\text{pt-s})\right),\\
    \end{split}
\end{equation}
where $\sigma$ and $\tau$ denote the softmax function and temperature scaling factor, respectively.

\noindent \textbf{Auxiliary Class Expansion.}
As described in \cref{subsec:3.1}, the text features for each class serve as anchors.
Here, logits can be viewed as a relative representation~\cite{relrep} of the image, where each logit element represents the relative distance between the image feature and the text features of corresponding anchor classes. 
As emphasized in \cite{relrep}, employing a sufficiently large number of anchors is crucial to facilitate effective transfer.

 Building on this insight, we incorporate additional auxiliary classes $C_\text{aux}$ alongside the task classes $C_\text{task}$ to construct the input logits. 
We collect the candidates of auxiliary classes from OpenImages~\cite{openimagev4}, which includes a broad range of real-world objects.
With the dimensionality of the logit space $M$, where $M >N_\text{task}$, we randomly sample auxiliary classes in a total number of $N_\text{aux} = M - N_\text{task}$. 
We compute logits that include auxiliary classes and pass them through the prediction adapter, which is written as:
\begin{equation}
\label{eq:aux_logits}
    \begin{split}
        z_\text{pt-s}&= [\textbf{sim}(\mathcal{V}_\text{pt-s}(x), \mathcal{T}_\text{pt-s}(c))]_{c\in C_\text{task}\cup C_\text{aux}}\in\R^{M}.\\
    \end{split}
\end{equation}
Note that the transition matrix $W_{\text{s}}$ is re-shaped to $\mathbb{R}^{D \times M}$, where $D$ refers to the dimensionality of the latent space.
 With the reshaped transition matrix $W_{\text{s}}$, the output logits will have a dimensionality of $M$.
Here, only the output logits corresponding to the task classes are used.

\subsection{Forward-only Adapter Transfer}
\label{sec:3.3}
The primary goal of TransMiter is to transfer adaptation knowledge acquired from weaker to stronger models without fine-tuning it directly.
A straightforward transfer is directly using the logits from the pre-trained strong VLM $z_\text{pt-t}$ as TransMiter’s input, which can be written as:
\begin{equation}
\label{eq:upa_target}
    \begin{split}
        z_\text{pt-t}&=[\textbf{sim}(\mathcal{V}_\text{pt-t}(x),\mathcal{T}_\text{pt-t}(c))]_{c\in C_\text{task}\cup C_\text{aux}}\in\R^{M},\\
        \hat{z}_\text{t} &= f(z_\text{pt-t}W_\text{s}^\intercal)W_\text{s}\in\R^{M},\\
        \hat{z}_\text{t}&\mathrel{\mathop:}=[\hat{z}_\text{t}(c)]_{c\in C_\text{task}}\in\R^{N_\text{task}},
    \end{split}
\end{equation}
where $\mathcal{V}_\text{pt-t}$ and $\mathcal{T}_\text{pt-t}$ denote the image and text encoder of pre-trained strong VLM, respectively.
To maintain semantic consistency, the same classes ($C_\text{task}\cup C_\text{aux}$) must be used to compute the logits for the pre-trained strong VLM as those used for weak VLM.

 However, since our adapter is initially trained on the distribution of weak VLM’s logits, discrepancies between the logit distributions of weak and strong VLMs lead to suboptimal transferability.
To address this, we focus on aligning the input features $h_\text{s}$ and $h_\text{t}$ from weak and strong VLMs, resulting in strong VLM's transition matrix, \ie basis.

\begin{table*}[!t]
\centering
\footnotesize
\setlength{\tabcolsep}{3pt}
\resizebox{\linewidth}{!}{
\begin{tabular}{lcc |c|c|cccccccccccc}
\toprule

\ \ \ \textbf{Source $\rightarrow$ Target}
&  \multirow{2}{*}{\textbf{Model} }
& \multirow{2}{*}{\textbf{Method}}
& \textbf{FLOPs}
& \multirow{2}{*}{\textbf{Avg.}}
& \textbf{Image}
& \multirow{2}{*}{\textbf{Caltech} }
& \multirow{2}{*}{\textbf{Pets} }
& \multirow{2}{*}{\textbf{Cars}}
& \multirow{2}{*}{\textbf{Flowers}}
& \multirow{2}{*}{\textbf{Food} }
& \textbf{Air}
& \multirow{2}{*}{\textbf{SUN}}
& \multirow{2}{*}{\textbf{DTD} }
& \textbf{Euro}
& \multirow{2}{*}{\textbf{UCF}} \\
\multicolumn{1}{c}{\textbf{(Strategy)}} &
 &&
\textbf{(G)}&&
\textbf{Net} &&&&&&
\textbf{craft}&&&
\textbf{SAT} \\
\toprule
\midrule
\rowcolor{cyan!10}\multirow{7}{*}{\shortstack{{RN50 $\rightarrow$ ViT-B/16}\\(CoOP)}} 
\cellcolor{white}& \cellcolor{white}{Source} &Fine-tuning& 0&73.39 & 62.91 & 91.56 & 86.25 & 73.20 & \underline{94.74} & 74.57 & 32.05 & 68.96 & 63.44 & \underline{83.33} & 76.32\\
\cmidrule{2-16}

\rowcolor{cyan!10}
\cellcolor{white}& \cellcolor{white}{Target}& {Zero-shot} & 0&65.33 & 66.72 & 93.31 & \underline{89.10} & 65.54 & 70.77 & \textbf{85.88} & 24.81 & 62.57 & 44.09 & 48.38 & 67.46\\
& &Prompt Transfer&0 &62.74 & 64.27 & 90.63 & 86.43 & 59.03 & 63.37 & \underline{85.00} & 4.13 & 55.63 & 38.10 & 78.40 & 65.10\\
& &EFT & 12.27& \underline{75.78} & \underline{66.55} & \underline{94.32} & {88.51} & \underline{74.64} & 94.07 & 81.63 & \textbf{33.90} & \underline{70.60} & \underline{64.30} & \textbf{85.82} & \underline{79.19}
\\
\rowcolor{red!10}
\cellcolor{white}&\cellcolor{white} &\textbf{TransMiter} (ours) &0.01& \textbf{77.16} & \textbf{69.68} & \textbf{95.21} & \textbf{91.60} & \textbf{75.92} & \textbf{95.13} & 84.52 & \underline{33.69} & \textbf{72.76} & \textbf{67.38} & 82.62 & \textbf{80.30} \\
\rowcolor{gray!20}
\cellcolor{white}
&\cellcolor{white}&\textcolor{gray}{Fine-tuning} &\textcolor{gray}{0}& \textcolor{gray}{79.92} & \textcolor{gray}{71.86} & \textcolor{gray}{95.48} & \textcolor{gray}{91.94} & \textcolor{gray}{82.65} & \textcolor{gray}{97.37} & \textcolor{gray}{84.22} & \textcolor{gray}{43.46} & \textcolor{gray}{74.86} & \textcolor{gray}{68.68} & \textcolor{gray}{85.44} & \textcolor{gray}{83.12}\\

\midrule
\rowcolor{cyan!10}
\multirow{7}{*}{\shortstack{RN50$\rightarrow$ ViT-L/14 \\ (CoOP)}} 
\cellcolor{white}&\cellcolor{white}{Source} &Fine-tuning &0&73.39 & 62.91 & 91.56 & 86.25 & 73.20 & 94.74 & 74.57 & 32.05 & 68.96 & 63.44 & 83.33 & 76.32 \\
\cmidrule{2-16}
\rowcolor{cyan!10}
\cellcolor{white}& \cellcolor{white}{Target} &Zero-shot &0& 72.54 & 73.46 & 95.13 & \underline{93.49} & 76.88 & 79.46 & \underline{90.91} & 32.55 & 67.66 & 53.07 & 60.33 & 74.99\\
& &Prompt Transfer &0& 77.28 & \underline{75.00} & \underline{96.07} & \textbf{94.00} & 75.83 & 85.70 & \textbf{90.97} & 32.17 & 72.60 & 61.20 & \underline{87.03} & 79.50\\
& &EFT &12.27& \underline{80.27} & 73.99 & 96.05 & 91.31 & \textbf{82.12} & \underline{95.70} & 87.91 & \textbf{40.01} & \underline{73.99} & \underline{68.75} & \textbf{89.24} & \textbf{83.89} \\
\rowcolor{red!10}
\cellcolor{white}&\cellcolor{white}&\textbf{TransMiter} (ours) &0.01& \textbf{80.39} & \textbf{75.60} & \textbf{96.52} & 93.42 & \underline{81.39} & \textbf{96.06} & 89.50 & \underline{39.52} & \textbf{75.21} & \textbf{69.90} & 83.77 & \underline{83.36}
\\
\rowcolor{gray!20}
\cellcolor{white}
&\cellcolor{white}&\textcolor{gray}{Fine-tuning}& 
\textcolor{gray}{0}& \textcolor{gray}{84.57} & \textcolor{gray}{78.24} & \textcolor{gray}{97.00} & \textcolor{gray}{94.41} & \textcolor{gray}{88.88} & \textcolor{gray}{98.98} & \textcolor{gray}{89.79} & \textcolor{gray}{56.28} & \textcolor{gray}{77.76} & \textcolor{gray}{72.97} & \textcolor{gray}{88.57} & \textcolor{gray}{87.39}\\
\midrule
\rowcolor{cyan!10}
\multirow{7}{*}{\shortstack{ViT-B/16$\rightarrow$ViT-L/14 \\ (CoOP)}} 
\cellcolor{white}&\cellcolor{white}{Source} &Fine-tuning&0&79.92 & 71.86 & 95.48 & 91.94 & 82.65 & \textbf{97.37} & 84.22 & \underline{43.46} & 74.86 & 68.68 & 85.44 & 83.12\\
\cmidrule{2-16}
\rowcolor{cyan!10}
\cellcolor{white}& \cellcolor{white}{Target} &Zero-shot & 0&72.54 & 73.46 & 95.13 & 93.49 & 76.88 & 79.46 & \underline{90.91} & 32.55 & 67.66 & 53.07 & 60.33 & 74.99 \\
& &Prompt Transfer &0& 78.50 & \underline{76.77} & 96.27 & \textbf{94.83} & 78.27 & 86.37 & \textbf{91.30} & 36.33 & 73.73 & 62.40 & \textbf{87.27} & 80.00
\\
& &EFT &35.16& \underline{81.04} & 74.90 & \underline{96.59} & 93.20 & \textbf{84.47} & 96.14 & 88.49 & 41.93 & \underline{75.01} & \underline{69.98} & \underline{86.17} & \underline{84.51}\\
\rowcolor{red!10}
\cellcolor{white}&\cellcolor{white}&\textbf{TransMiter} (ours) &0.01&\textbf{82.05} & \textbf{76.78} & \textbf{96.96} & \underline{94.24} & \underline{84.27} & \underline{97.25} & 90.15 & \textbf{44.38} & \textbf{76.72} &\textbf{72.04} & 84.72 & \textbf{85.01}\\
\rowcolor{gray!20}
\cellcolor{white}
&\cellcolor{white}
&\textcolor{gray}{Fine-tuning} &\textcolor{gray}{0}& \textcolor{gray}{84.57} & \textcolor{gray}{78.24} & \textcolor{gray}{97.00} & \textcolor{gray}{94.41} & \textcolor{gray}{88.88} & \textcolor{gray}{98.98} & \textcolor{gray}{89.79} & \textcolor{gray}{56.28} & \textcolor{gray}{77.76} & \textcolor{gray}{72.97} & \textcolor{gray}{88.57} & \textcolor{gray}{87.39}\\

\midrule
\rowcolor{cyan!10}
\multirow{7}{*}{\shortstack{ViT-B/16 $\rightarrow$ ViT-L/14 \\ (PromptSRC)}} 
\cellcolor{white}&\cellcolor{white}{Source} &Fine-tuning&0.37&\underline{81.74} & 72.81 & 95.77 & 93.85 & 80.89 & 97.04 & 87.50 & \textbf{45.20} & \underline{76.73} & \underline{73.31} & \textbf{91.02} & \underline{85.05}\\
\cmidrule{2-16}
\rowcolor{cyan!10}
\cellcolor{white}& \cellcolor{white}{Target}& {Zero-shot} & 0&72.54 & 73.46 & 95.13 & 93.49 & 76.88 & 79.46 & \underline{90.91} & 32.55 & 67.66 & 53.07 & 60.33 & 74.99 \\
& &Prompt Transfer&1.34 & 73.41 & 73.23& 92.73  &  91.43 & 75.03 & 80.57  & 90.00 & 30.37 & 68.77 & 54.27 
&75.90 & 75.20  \\
& &EFT & 36.84& 81.34 & \underline{75.98} & \underline{96.09} & \underline{94.22} & \underline{82.80} & \underline{95.51} & 90.85 & \underline{41.56} & 75.20 & 72.58 & 85.64 & 84.26\\

\rowcolor{red!10}
\cellcolor{white}&\cellcolor{white} &\textbf{TransMiter} (ours) &0.01&\textbf{82.51} & \textbf{77.11} & \textbf{97.03} & \textbf{94.93} & \textbf{83.11} & \textbf{97.39} & \textbf{91.66} & \underline{42.71} & \textbf{77.14} & \textbf{74.09} & \underline{86.95} & \textbf{85.52}
 \\
\rowcolor{gray!20}
\cellcolor{white}
&\cellcolor{white}
&\textcolor{gray}{Fine-tuning} &\textcolor{gray}{1.31}& \textcolor{gray}{85.17} & \textcolor{gray}{79.05} & \textcolor{gray}{97.38} & \textcolor{gray}{95.07} & \textcolor{gray}{86.85} & \textcolor{gray}{98.66} & \textcolor{gray}{91.96} & \textcolor{gray}{53.07} & \textcolor{gray}{79.97} & \textcolor{gray}{75.73} & \textcolor{gray}{90.92} & \textcolor{gray}{88.23}\\

\bottomrule
\end{tabular}}
\caption{\textbf{Performance on base-to-base adaptation transfer.} We combine source and target models among RN50, ViT-B/16, and ViT-L/14.
Gray-colored rows represent the performance of the target model when fine-tuned, serving as the upper bound.
}
\label{tab:fewshot}
\end{table*}

\noindent \textbf{Basis Change.}
 After training TransMiter, unlabeled images $x\in\mathcal{D}_\text{u}$ are used to extract the input feature of weak VLM $h_\text{s}$ by multiplying $z_\text{pt-s}$ with $W_\text{s}^\intercal$.
The same images are used to compute strong VLM logits $z_\text{pt-t}$ and obtain the input feature of strong VLM $h_\text{t}$ by also multiplying $z_\text{pt-t}$ with $W_\text{s}^\intercal$. 
Our goal is to find $\hat{W}$, a mapping between the input feature of weak and strong VLMs, by minimizing the distance between $h_\text{s}$ and $h_\text{t}$ of all the images.
With a regularization term, the objective can be written as:
\begin{equation}
    \begin{split}
        &H_\text{s} = [z_\text{pt-s}\left(C_\text{task}\cup C_\text{aux};x \right)W_\text{s}^\intercal]_{x\in\mathcal{D}_\text{u}}\in\R^{|\mathcal{D}_\text{u}|\times D}, \\
        &H_\text{t} = [z_\text{pt-t}\left(C_\text{task}\cup C_\text{aux};x \right)W_\text{s}^\intercal]_{x\in\mathcal{D}_\text{u}}\in\R^{|\mathcal{D}_\text{u}|\times D},\\
        &\hat{W}=\argmin_{W\in\mathcal{O}}||H_\text{t} W-H_\text{s}||_2^2+\beta||W-I||_2^2,\\   
    \end{split}
\end{equation}
where $\mathcal{O}=\{W\in\R^{D\times D}, W^\intercal W=I_{D}\}$ denotes the orthogonality constraint.
The mapping matrix $\hat{W}$ can be derived as the solution to the Orthogonal Procrustes problem~\citep{procrutes} via singular value decomposition, making backpropagation unnecessary.
The solution can be written as:
\begin{equation}
    \begin{split}
        U,S,V &= \textbf{SVD}(H_\text{t}^\intercal H_\text{s}+\beta I),\\
        \hat{W}&=UV^\intercal,\\
    \end{split}
\end{equation}
where $\beta$ refers to the regularization weight.

Here, the projection matrix for strong VLM is updated to $W_\text{s}^\intercal\hat{W}$.
Likewise, the reconstruction matrix for strong VLM is updated to $\hat{W}^\intercal W_\text{s}$. 
By doing so, the inverse relationship between the projection and reconstruction matrices and their orthogonality are preserved.
This process ensures that strong VLM retains its pre-trained knowledge while absorbing adaptation knowledge from weak VLM.
The final outputs of the enhanced strong VLM can be written as:
\begin{equation}
\label{eq:upa_target_final}
    \begin{split}
        \hat{z}_\text{t} &= f(z_\text{pt-t}W_\text{t}^\intercal )W_\text{t}\in\R^{M},\\
        \hat{z}_\text{t}&\mathrel{\mathop:}=[\hat{z}_\text{t}(c)]_{c\in C_\text{task}}\in\R^{N_\text{task}}.\\
    \end{split}
\end{equation}


\section{Experiments}
\subsection{Experimental setup}
\label{exp:setup}
\textbf{Settings.}
During the adaptation knowledge transfer, we need a fine-tuned weak VLM.
Based on how this model was fine-tuned, we set up two main evaluation setups: (1) \textbf{Base-to-base adaptation transfer}, where the weak model is trained on all the task classes, and (2) \textbf{Base-to-novel adaptation transfer}, where the weak model is fine-tuned on a subset of classes (\ie seen classes), and tested on both seen and the remaining classes (\ie unseen classes).
Through these configurations, we aim to evaluate how effectively the methods transfer task-specific adaptation knowledge from weaker to stronger models while preserving their generalized knowledge.
We evaluate our method on 11 visual recognition datasets for both settings.
See Supplement for more details.


\noindent \textbf{Baselines.}
\sethlcolor{cyan!10}
We adopt CLIP \cite{clip} for VLMs, varying with architecture and size, \eg ResNet-50 (RN50), ViT-B/16, and ViT-L/14.
We assume that the target model has stronger generalization capabilities than the source model in terms of its size and architecture.
For fine-tuning the source model, we apply CoOp~\cite{zhou2022coop} and PromptSRC~\cite{promptsrc}.
In each setting, we evaluate three methods within our framework: (1) \hl{\textbf{Source Fine-tuning}} $\theta_\text{ft-s}$, where the source model is fine-tuned; (2) \hl{\textbf{Target Zero-shot}} $\theta_\text{pt-t}$, the zero-shot performance of the pre-trained target model; and 
\sethlcolor{red!10} 
(3) \hl{\textbf{Target TransMiter}} $\theta_\text{t}^*$, where TransMiter is applied to the pre-trained target model.
\sethlcolor{cyan!10}
As described in \cref{subsec:3.1}, our objective is to improve the strong model beyond the performance of the fine-tuned weak model (\hl{\textbf{Source Fine-tuning}}) and pre-trained strong model (\hl{\textbf{Target Zero-shot}}).
\sethlcolor{gray!20}
Additionally, we report the performance of \hl{\textbf{Target Fine-tuning}} \textcolor{gray}{$\theta_\text{ft-t}$}, where the strong model is fine-tuned.
It serves as the `upper bound'.

 To compare with transferable adaptation methods, we employ the following baselines: \textbf{EFT}~\cite{Emulator,proxy_tuning} and \textbf{Prompt Transfer}~\cite{prompt_trans}.
Prompt Transfer transfers learned soft prompts from weaker to stronger models.
This approach requires resource-intensive training to align the representation spaces between weak and strong models. 
To achieve this, we adopt a naive knowledge distillation method~\cite{li2024promptkd, naive_kd}.
In contrast, EFT operates in a training-free manner but relies on multiple weak models to assist the strong model during inference, leading to high inference costs.
Additional details are provided in the Supplement.

{\begin{table}[!t]
  \centering
  \footnotesize
  \setlength{\tabcolsep}{5pt}
  \resizebox{\linewidth}{!}{
    \begin{tabular}{lc>{\columncolor{cyan!10}}c|>{\columncolor{cyan!10}}ccc>{\columncolor{red!10}}c>{\columncolor{gray!20}}c}
        \toprule
        \multirow{3}{*}{\textbf{Dataset}} & \multirow{3}{*}{\textbf{Set}} & \cellcolor{white}{\textbf{Source}}  & \multicolumn{5}{c}{\textbf{Target}}\\
        \cmidrule{3-8}
        & & \cellcolor{white}{Fine-} & \cellcolor{white}{Zero-}& Prompt-& \multirow{2}{*}{EFT} & \cellcolor{white}{\textbf{TransMiter}} & \cellcolor{white}\textcolor{gray}{Fine-} \\
        & &  \cellcolor{white}{tuning} & \cellcolor{white}{shot }& transfer & & \cellcolor{white}(ours) & \cellcolor{white} \textcolor{gray}{tuning} \\
        \midrule
        \midrule
        
        \multirow{3}{*}{\shortstack[l]{Average on \\11 datasets}}  & Base &84.19 & 76.71 &76.07  &\underline{84.32} &   \textbf{85.27} & \textcolor{gray}{87.15} \\
        & Novel & 75.35 & \underline{80.85} &77.84 &80.24 & \textbf{81.07} &\textcolor{gray}{81.59} \\
        & HM & 79.53 & 78.73 &76.94 &\underline{82.23} & \textbf{83.12} &\textcolor{gray}{84.28} \\
        \midrule

        \multirow{3}{*}{ImageNet} & Base & 77.73 & 79.19 & 78.03 & \underline{80.83} & \textbf{81.73} & \textcolor{gray}{83.05} \\
        & Novel & 70.42 & 74.03 & 73.83 & \underline{74.43} & \textbf{75.84} & \textcolor{gray}{76.97} \\
        & HM & 73.89 & 76.52 & 75.87 & \underline{77.50} & \textbf{78.67} & \textcolor{gray}{79.90} \\
        \midrule
         \multirow{3}{*}{Caltech} & Base & \underline{98.19} & 95.61 & 92.97 & 96.82 & \textbf{98.58} & \textcolor{gray}{98.47} \\
         & Novel & 94.21 & \textbf{97.71} & 95.47 & \underline{96.72} & 96.29 & \textcolor{gray}{97.42} \\
         & HM & 96.16 & 96.65 & 94.20 & \underline{96.77} & \textbf{97.42} & \textcolor{gray}{97.94} \\
        \midrule
        \multirow{3}{*}{Pets} & Base & \underline{95.39} & 95.16 & 92.93 & 94.52 & \textbf{95.80} & \textcolor{gray}{96.01} \\
         & Novel & {95.41} & \textbf{98.10} & \underline{97.37} & 97.13 & 97.20 & \textcolor{gray}{98.68} \\
         & HM & 95.40 & \textbf{96.61} & 95.10 & 95.81 & \underline{96.50} & \textcolor{gray}{97.33} \\
        \midrule
        \multirow{3}{*}{Cars} & Base & 78.29 & 74.51 & 71.60 & \underline{79.47} & \textbf{79.78} & \textcolor{gray}{84.06} \\
         & Novel & 75.13 & \textbf{84.67} & 83.77 & 83.88 & \underline{84.01} & \textcolor{gray}{84.79} \\
         & HM & 76.68 & 79.27 & 77.21 & \underline{81.62} & \textbf{81.84} & \textcolor{gray}{84.42} \\
        \midrule
         \multirow{3}{*}{Flowers} & Base & \textbf{97.94} & 82.72 & 81.10 & 95.16 & \underline{97.28} & \textcolor{gray}{98.73} \\
         & Novel & 76.93 & \textbf{82.98} & 78.77 & 81.61 & \underline{82.01} & \textcolor{gray}{82.48} \\
         & HM & 86.17 & 82.85 & 79.92 & \underline{87.86} & \textbf{88.99} & \textcolor{gray}{89.88} \\
        \midrule
        \multirow{3}{*}{Food101} & Base & 90.68 & \underline{93.75} & 92.90 & 93.71 & \textbf{94.18} & \textcolor{gray}{94.29} \\
         & Novel & 91.20 & \textbf{94.92} & 93.33 & 93.97 & \underline{94.63} & \textcolor{gray}{95.04} \\
         & HM & 90.94 & \underline{94.33} & 93.11 & 93.84 & \textbf{94.40} & \textcolor{gray}{94.66} \\
        \midrule
          \multirow{3}{*}{Aircraft} & Base & 42.84 & 37.21 & 34.47 & \underline{44.68} & \textbf{44.78} & \textcolor{gray}{52.32} \\
         & Novel & 38.11 & \textbf{44.33} & 39.07 & \underline{44.01} & 43.39 & \textcolor{gray}{43.55} \\
         & HM & {40.34} & 40.46 & 36.63 & \textbf{44.34} & \underline{44.07} & \textcolor{gray}{47.54} \\
        \midrule
        \multirow{3}{*}{SUN397} & Base & \underline{82.67} & 73.52 & 74.20 & 81.42 & \textbf{82.69} & \textcolor{gray}{85.05} \\
         & Novel & \underline{78.54}& 77.72 & 77.50 & 78.44 & \textbf{80.85} & \textcolor{gray}{81.42} \\
         & HM & \underline{80.55} & 75.56 & 75.81 & {79.90} & \textbf{81.76} & \textcolor{gray}{83.19} \\
        \midrule
        \multirow{3}{*}{DTD} & Base & \textbf{83.37} & 61.23 & 57.63 & \underline{82.02} & \textbf{83.37} & \textcolor{gray}{84.57} \\
         & Novel & 58.53 & \textbf{70.89} & 59.10 & 68.68 & \underline{69.08} & \textcolor{gray}{70.41} \\
         & HM & 68.78 & 65.71 & 58.36 & \underline{74.76} & \textbf{75.56} & \textcolor{gray}{76.84} \\
        \midrule
        \multirow{3}{*}{EuroSAT} & Base & \underline{92.26} & 70.93 & 83.13 & 92.19 & \textbf{92.94} & \textcolor{gray}{93.29} \\
         & Novel & 71.71 & 83.44 & 77.80 & \underline{82.69} & \textbf{85.82} & \textcolor{gray}{83.71} \\
         & HM & 80.70 & 76.68 & 80.38 & \underline{87.18} & \textbf{89.24} & \textcolor{gray}{88.24} \\
        \midrule
        \multirow{3}{*}{UCF} & Base & \underline{86.73} & 79.94 & 77.83 & 86.66 & \textbf{86.88} & \textcolor{gray}{88.81} \\
         & Novel & 78.64 & {80.58} & 80.27 & \underline{81.11} & \textbf{82.64} & \textcolor{gray}{83.07} \\
         & HM & 82.48 & 80.26 & 79.03 & \underline{83.79} & \textbf{84.71} & \textcolor{gray}{85.85} \\
    \bottomrule
    \end{tabular}}
\caption{\textbf{Performance on base-to-novel adaptation transfer.} 
Source and target models are ViT-B/16 and ViT-L/14, respectively.
We adopt PromptSRC as a fine-tuning strategy.}
\label{tab:b2n}
\end{table} }

{\small\begin{table}[!t]
    \centering
    \setlength{\tabcolsep}{5pt}
    \centering
    \footnotesize
    \begin{tabular}{l c| c c c | c}
    \toprule
    \multirow{1}{*}{\textbf{Strategy}}&\multirow{1}{*}{\textbf{Method}} & \multirow{1}{*}{Base} & \multirow{1}{*}{Novel} & \multirow{1}{*}{HM} & \multirow{1}{*}{ FPS}\\
    
    \midrule
    \midrule
    CLIP & \cellcolor{gray!20}Zero-shot & \cellcolor{gray!20}76.71 & \cellcolor{gray!20}80.85 & \cellcolor{gray!20}78.73 & \cellcolor{gray!20}286 \\
    CLIP & \cellcolor{red!10}\textbf{TransMiter+} & \cellcolor{red!10}86.00 & \cellcolor{red!10}80.77 & \cellcolor{red!10}83.30 & \cellcolor{red!10}285 \\
    \midrule
    MaPLe & \cellcolor{gray!20}Fine-tuning & \cellcolor{gray!20}85.58 & \cellcolor{gray!20}{79.35} & \cellcolor{gray!20}{82.35} & \cellcolor{gray!20}{261} \\
    MaPLe & \cellcolor{red!10}\textbf{TransMiter+} & \cellcolor{red!10}\textbf{87.41} & \cellcolor{red!10}\textbf{81.54} & \cellcolor{red!10}\textbf{84.37} & \cellcolor{red!10}\textbf{285} \\
    \midrule
    PromptSRC & \cellcolor{gray!20}Fine-tuning & \cellcolor{gray!20}87.15 & \cellcolor{gray!20}81.59 & \cellcolor{gray!20}84.28 & \cellcolor{gray!20}258 \\
    PromptSRC & \cellcolor{red!10}\textbf{TransMiter+} & \cellcolor{red!10}\textbf{87.59} & \cellcolor{red!10}\textbf{81.97} & \cellcolor{red!10}\textbf{84.69} & \cellcolor{red!10}\textbf{285} \\
    \midrule
    HPT & \cellcolor{gray!20}Fine-tuning & \cellcolor{gray!20}\textbf{87.98} & \cellcolor{gray!20}{82.13} & \cellcolor{gray!20}{84.95} & \cellcolor{gray!20}{250} \\
    HPT & \cellcolor{red!10}\textbf{TransMiter+} & \cellcolor{red!10}{87.53} & \cellcolor{red!10}\textbf{82.65} & \cellcolor{red!10}\textbf{85.02} & \cellcolor{red!10}\textbf{285} \\
    \midrule
    LwEIB & \cellcolor{gray!20}Fine-tuning & \cellcolor{gray!20}86.86 & \cellcolor{gray!20}\textbf{82.74} & \cellcolor{gray!20}84.75 & \cellcolor{gray!20}{148} \\
    LwEIB & \cellcolor{red!10}\textbf{TransMiter+} & \cellcolor{red!10}\textbf{87.55} & \cellcolor{red!10}{82.52} & \cellcolor{red!10}\textbf{84.96} & \cellcolor{red!10}\textbf{285} \\
    \bottomrule
    
\end{tabular}
    \caption{\textbf{Performance on supervised fine-tuning.}
    All the methods use ViT-L/14 as a base model. 
    TransMiter+ uses ViT-B/16 with each specified strategy for adaptation knowledge extraction.
        }
    \label{table:sota_comparison}
\end{table}}

\subsection{Experimental Results}
\label{exp:res}
\textbf{Base-to-base adaptation transfer.}
\cref{tab:fewshot} reports the performance of our method and baselines across 11 visual recognition datasets under the base-to-base adaptation transfer setting. 
To evaluate the inference efficiency of each method, we also reported the additionally introduced GFLOPs compared to the base model.

In all source-target combinations, TransMiter outperforms Target Zero-shot and Source Fine-tuning in average performance across 11 tasks.
With ResNet-50 as the source and ViT-B/16 as the target, TransMiter surpasses Target Zero-shot and source Fine-tuning by a margin of 11.77\% and 3.77\%. 
As the target model advances, for instance from ViT-B/16 to ViT-L/14 with ResNet-50 as the source, the performance increases further, from 77.16\% to 80.39\%.
This result is not surprising, as stronger zero-shot performance generally corresponds to enhanced fine-tuning capabilities.

Interestingly, using a stronger source model ViT-B/16 with ViT-L/14 as the target further improves the performance of the target model from 80.39\% to 82.05\%.
Furthermore, when a more advanced fine-tuning approach is applied to the source model, CoOp to PromptSRC, the performance of the target model with TransMiter increases from 82.05\% to 82.51\%.
These results demonstrate that more generalized and well-adapted models possess strong capabilities for transferring adaptation knowledge.

When comparing TransMiter with other transferable adaptation methods, it shows the best average accuracy across all source-target combinations.
This is especially notable given that Prompt Transfer requires substantial retraining efforts whenever the new model comes, while TransMiter uses a simple forward-only algorithm to update the adapter. 
Additionally, TransMiter achieves efficient adaptation with minimal computational overhead, requiring only 0.01 GFLOPs, while EFT incurs a significantly higher inference cost.
This indicates that TransMiter is not only effective in transferring adaptation knowledge but also efficient.

\noindent \textbf{Base-to-novel adaptation transfer.}
\cref{tab:b2n} presents an experiment under the base-to-novel adaptation transfer setting to verify whether the transferable adaptation methods can maintain generalization ability in transferred models.
On average performance across all datasets, TransMiter outperforms all baseline methods in all metrics: Base, Novel, and HM. 
Notably, TransMiter is the only method that maintains its Novel performance without any degradation compared to Target Zero-shot.
These results underscore TransMiter's superior ability to transfer specialized knowledge while maintaining strong generalization capabilities.

In addition to the unsupervised adaptation transfer setting, we further fine-tune TransMiter with labeled data (TransMiter+) to fully assess its capability, and compare it against recent state-of-the-art supervised fine-tuning approaches in \cref{table:sota_comparison}.
We integrate additional adaptation strategies, such as MaPLe~\cite{khattakMaPLe}, HPT~\cite{hpt}, and LwEIB~\cite{lweib}, into TransMiter for a weaker model (\eg ViT-B/16) adaptation, which serves as a `fine-tuned weak VLM' during the adaptation knowledge extraction process. 
Even without the adaptation knowledge extraction phase (row 2), TransMiter+ leads to a substantial improvement over the zero-shot baseline. 
Moreover, since TransMiter's knowledge extraction and transfer provide a strong initialization, fine-tuning TransMiter with supervision after the adaptation knowledge transfer consistently yields improved performance across various strategies. 
Importantly, combining any fine-tuning strategy with TransMiter not only surpasses the performance of its fine-tuning counterpart but also lowers inference cost, demonstrating both high effectiveness and efficiency.

\subsection{Ablation study}
\label{exp:abl}

\begin{table}[!t]
    \centering
    \setlength{\tabcolsep}{5pt}
    \resizebox{\linewidth}{!}{
    \footnotesize
    \begin{tabular}{c c c c c | c c c c}
        \toprule
        \multirow{2}{*}{\textbf{Model}} &  & \multicolumn{3}{c}{\textbf{Method}} & \multicolumn{3}{c}{\textbf{Base-to-novel}} & \multirow{2}{*}{\textbf{Base-to-base}} \\
        \cmidrule(r){3-5} \cmidrule(r){6-8}
         & & PA & ACE & BC & Base & Novel & HM & \\
        \midrule
        \midrule
        \cellcolor{white}{\multirow{4}{*}{Source} }
        & (a)& \multicolumn{3}{c|}{Zero-shot}  &   69.51 &74.29 & 71.72 & 65.33 \\
        & (b) &\checkmark &  & &  82.45 & 75.90 & 78.86& 79.30 \\
        & (c) & \checkmark & \checkmark & &  83.83 & 76.21 &79.62& 81.39\\
        \rowcolor{cyan!10}
        \cellcolor{white}& \textcolor{black}{(d)}&\multicolumn{3}{c|}{\textcolor{black}{Fine-tuning}}& \textcolor{black}{84.19} & \textcolor{black}{75.35} & \textcolor{black}{79.28} & \textcolor{black}{81.74} \\

        \midrule

        \rowcolor{cyan!10}
        \cellcolor{white}{\multirow{6}{*}{Target} }
        & (e)& \multicolumn{3}{c|}{Zero-shot} &  76.71 & 80.85 & 78.63& 72.54\\
        & (f)&\checkmark &  &   & 80.82 & 79.64 & 80.09&76.25 \\
        & (g)&\checkmark & \checkmark &  & 82.90 & 79.80 & 81.24&78.98\\
        & (h)&\checkmark &  & \checkmark & 81.60 & 80.24 & 80.80&77.65 \\
        \rowcolor{red!10}
        \cellcolor{white}& (i)&\checkmark & \checkmark & \checkmark  & \textbf{85.27} & \textbf{81.07} & \textbf{83.12} & \textbf{82.51}\\
        \bottomrule
    \end{tabular}}
    \caption{\textbf{Ablation study of TransMiter.} PA, ACE, and BC refer to ``\underline{P}rediction-based \underline{A}dapter'', ``\underline{A}uxiliary \underline{C}lass \underline{E}xpansion'', and ``\underline{B}asis \underline{C}hange'', respectively.}
    \label{table:comp_upa}
\end{table}
In this section, we evaluate the effectiveness of TransMiter through ablations. 
Unless otherwise stated, we use ViT-B/16 and ViT-L/14 as the source and target VLMs, respectively, and PromptSRC~\cite{promptsrc} for fine-tuning.

\noindent \textbf{Component analysis in TransMiter.}
\cref{table:comp_upa} provides a comparison of TransMiter components when applying to both source and target models, with results reported as the average accuracy of all tasks.
When the prediction-based adapter (PA) is applied naively to the source model ((a)$\rightarrow$(b)), it significantly improves the source model's zero-shot performance, but still lags behind the performance of the fine-tuned source model ((b)$\mathrm{\ vs. \ }$(d)). 
With the addition of auxiliary classes ((b)$\rightarrow$(c)), the prediction adapter better approximates or even surpasses the fine-tuned source model in certain cases ((c)$\mathrm{\ vs. \ }$(d)), such as the performance of Novel and HM in the base-to-novel setting. 
These results indicate that TransMiter can effectively capture the adaptation knowledge from the source model.

Also, naively applying the adapter to the target model ((e)$\rightarrow$(f)) leads to an improvement over the zero-shot target model.
When auxiliary classes (ACE) are introduced ((f)$\rightarrow$(g)), the target model is further enhanced.
Additionally, incorporating basis change (BC) with the adapter ((f)$\rightarrow$(h)) also contributes to better transferability.
When both methods are applied together in the target model ((i)), it achieves the highest performance among the tested configurations. 
These results demonstrate that using auxiliary classes and basis change methodology has strong complementarity and serves as a highly effective method for enhancing the transferability of TransMiter.

\begin{table}[t]
    \centering
    \setlength{\tabcolsep}{5pt}
    \resizebox{\linewidth}{!}{
    \footnotesize
    \begin{tabular}{c c c c c c}
        \toprule
        \multirow{2}{*}{\textbf{Method}}
        &\multirow{2}{*}{\textbf{Labels}}
        & \multicolumn{2}{c}{\textbf{Transfer}}
        & \multicolumn{1}{c}{\textbf{Inference}}
        & \multirow{2}{*}{\textbf{HM}}\\
          & & Time (s) & Mem. (Mb)  & FPS&  \\
        
        \midrule
        \midrule

        \rowcolor{cyan!10}
        EFT & \xmark& 0& 0 & 169 & 81.62\\

        \rowcolor{cyan!10}
        Prompt Transfer&\xmark& 480& 33188  & 258 & 77.21\\
        \rowcolor{red!10}
        \textbf{TransMiter} & \xmark&24 & 3784 & \textbf{285} &\textbf{81.84}\\
        
        \midrule
        \rowcolor{gray!10}
        PromptSRC &\checkmark& 321& 31014& 258& 84.42 \\
        \rowcolor{red!10}
        \textbf{TransMiter+}&\checkmark& 52 & 2996 & \textbf{285} &\textbf{85.16}\\

        \bottomrule
    \end{tabular}}
    \caption{\textbf{Efficiency analysis.} 
    Computation costs are measured using a single A6000 GPU on StanfordCars dataset.
    }
    \label{tab:efficiency_analysis}
\end{table}

\noindent \textbf{Efficiency analysis.}
In \cref{tab:efficiency_analysis}, we evaluate the computational costs of TransMiter on StanfordCars dataset.
Among unsupervised transfer methods, TransMiter achieves strong performance and fast inference, while its backpropagation-free design significantly reduces transfer time and memory.
With a few labeled data, TransMiter+ surpasses the performance and inference speed of the fine-tuning strategy (PromptSRC) used in its development.
This improvement takes less than a minute of extra training, as it updates only the adapter on top of the VLM, whereas PromptSRC requires almost full-model gradient computation, highlighting TransMiter's superior efficiency.

\section{Conclusion}
We present TransMiter, a transferable model-agnostic adapter that enhances vision-language models without backpropagation. 
TransMiter utilizes the model’s prediction to capture adaptation knowledge, enabling seamless knowledge transfer from weaker to stronger models, regardless of architecture or size.
With its simple structure and forward-only transfer, TransMiter ensures high computational efficiency. 
Experimental results demonstrate its exceptional transferability across various models, offering a practical solution for scalable and efficient adaptation.

\section*{Acknowledgments}
This work was partly supported by Korea Research Institute for defense Technology planning and advancement - Grant funded by Defense Acquisition Program Administration(DAPA)(KRIT-CT-23-021) (30\%), the National Research Foundation of Korea (NRF) grant funded by the Korea government (MSIT) (NRF-2023R1A2C2005373) (25\%), Institute for Information \& communications Technology Promotion (IITP) grant funded by the Korea government (MSIP) (No. RS-2024-00443251, Accurate and Safe Multimodal, Multilingual Personalized AI Tutors) (25\%), and the Korea Health Technology R\&D Project through the Korea Health Industry Development Institute (KHIDI), funded by the Ministry of Health \& Welfare, Republic of Korea (grant number: HR20C0021) (20\%).

\bibliography{aaai2026}

\newpage

\appendix
\appendix

\section*{Appendix}
This supplement offers further details on the experimental settings, additional experimental results, computational costs, and the derivation of our closed-form solution.
This includes \textbf{(A)} experimental settings, \textbf{(B)} additional experiments, \textbf{(C)} additional ablations and analyses,
and \textbf{(D)} derivation of closed-form solution.

\section{Experimental settings}
\subsection{Settings}
We primarily report results for two adaptation transfer settings: base-to-base and base-to-novel.
For evaluation metrics, base-to-base adaptation transfer provides accuracy for all classes. Likewise, base-to-novel adaptation transfer reports the accuracy for the seen classes (Base), the unseen classes (Novel), and the harmonic mean (HM) of these seen and unseen class accuracies. The reported accuracy is the average accuracy over 3 seeds (\eg 1,2,3).

\subsection{Datasets}
We evaluate our method on 11 visual recognition datasets for both base-to-base and base-to-novel adaptation transfer settings, including ImageNet \cite{imagenet}, EuroSAT \cite{eurosat}, DTD \cite{dtd}, Food101 \cite{food}, OxfordPets \cite{pets}, FGVCAircraft \cite{aircraft}, Flowers102 \cite{flowers}, UCF101 \cite{ucf101}, Caltech101 \cite{caltech}, StanfordCars \cite{cars}, and SUN397 \cite{sun}. 
For fine-tuning the models, we follow CoOp \cite{zhou2022coop} and CoCoOp~\cite{zhou2022cocoop}, using a 16-shot setup. 
All unlabeled image samples~\cite{li2024promptkd} are used for training and transfer in TransMiter. 
We use all the unlabeled images in each dataset to train and transfer TransMiter.
Specifically, for base-to-novel adaptation transfer experiments, we adopt a transductive setting~\cite{pseudoclip,li2024promptkd}, where images from both seen and unseen classes are accessible during training. 
\cref{tab:stats} provides the data statistics.
\begin{table}[ht]
\centering
\footnotesize
\resizebox{\linewidth}{!}{
\begin{tabular}{l|ccc}
\toprule
\textbf{Dataset} & \textbf{\# Train }& \textbf{\# Classes} & \textbf{Manual prompt}\\
\midrule
\midrule
ImageNet & 1,281,167& 1000&"a photo of a \{$c$\}."\\
EuroSAT & 13,500&10&"a centered satellite photo of  \{$c$\}."\\
DTD & 2,820&47&"\{$c$\} texture."\\
Food101 & 50,500&101& "a photo of \{$c$\}, a type of food."\\
Pets & 2,944&37&"a photo of a \{$c$\}, a type of pet."\\
Aircraft &3,334 &100&"a photo of a \{$c$\}, a type of aircraft."\\
Flowers & 4,093&102&"a photo of a \{$c$\}, a type of flower."\\
UCF & 7,649&101& "a photo of a person doing \{$c$\}."\\
Caltech & 4,128&100&"a photo of a \{$c$\}."\\
Cars &6,509 &196&"a photo of a \{$c$\}."\\
SUN & 15,880&397&"a photo of a \{$c$\}."\\

\bottomrule
\end{tabular}
}
\caption{\textbf{Data statistics.} Manual prompt is used for pre-trained VLMs. $c$ denotes class name.}
\label{tab:stats}
\end{table}

\subsection{Baselines}
Since adaptation knowledge transfer has not been sufficiently studied in vision-language models, we adopt the transferable adaptation methods from the language domain for the baselines: EFT~\cite{Emulator} and Prompt Transfer~\cite{prompt_trans}.
Note that the baseline methods operate without the need for labeled data as in TransMiter.

\noindent \textbf{EFT} is a ``parameter-based transfer'' method that utilizes multiple weak models to assist strong models during inference by incorporating predicted logits.
Specifically, the difference between the logits of fine-tuned weak VLM $z_\text{ft-s}$ and pre-trained weak VLM $z_\text{pt-s}$ serve as adaptation knowledge, which is then added to the pre-trained strong VLM's logits $z_\text{pt-t}$ for the knowledge transfer.
The formulation of EFT can be written as:
\begin{equation}
    \begin{split}
        z_\text{pt-s}&=[\textbf{sim}(\mathcal{V}_\text{pt-s}(x),\mathcal{T}_\text{pt-s}(c))]_{c\in C_\text{task}},\\
        z_\text{ft-s}&=[\textbf{sim}(\mathcal{V}_\text{ft-s}(x),\mathcal{T}_\text{ft-s}(c))]_{c\in C_\text{task}},\\
        z_\text{pt-t}&=[\textbf{sim}(\mathcal{V}_\text{pt-t}(x),\mathcal{T}_\text{pt-t}(c))]_{c\in C_\text{task}},\\
        \hat{z}_\text{t}&=z_\text{pt-t}+\alpha(z_\text{ft-s}-z_\text{pt-s}),\\
    \end{split}
    \label{eq:eft}
\end{equation}
where $\alpha$ denotes the scaling factor.

EFT is a strong baseline, as it does not require additional training costs to transfer adaptation knowledge. 
However, it relies on three models, two weak VLMs (pre-trained and fine-tuned) and one strong VLM, resulting in additional inference costs.

\noindent 
\textbf{Prompt Transfer} is an approach that utilizes previously learned soft prompts from the fine-tuned weak VLM as adaptation knowledge and transfers them to strong VLM for prompt initialization.
As models differ in size and architectural design, dimensional mismatches and semantic inconsistencies \cite{LLMbb} often lead to poor or infeasible transferability. 
Consequently, additional training is necessary to enable the transfer of soft prompts. 
When the dimensionality of soft prompts between weak and strong VLMs differ, we employ a single projection matrix to align the dimensionality and only train this projection matrix while keeping the transferred soft prompts frozen.
Conversely, if the dimensionality is the same, we train only the transferred soft prompts.
Although adding a few soft prompts results in relatively small additional inference costs compared to EFT, it requires high transfer computation costs due to its backpropagation-based transfer (see Table 6 in the main).

\noindent 
\textbf{Fine-tuning} In our experiments, we adopt CoOp~\cite{zhou2022coop} and PromptSRC~\cite{promptsrc} as the methods for soft prompts learning during fine-tuning.
Specifically, when transferring CoOp~\cite{zhou2022coop} prompts, we learn a single projection matrix, as learnable prompts are only introduced in the first transformer block of the text encoder. 
However, since PromptSRC~\cite{promptsrc} uses deep prompting~\cite{khattakMaPLe}, which learns separate prompts at different transformer blocks of the image and text encoder, we employ separate projection matrices for each layer and encoder.

To ensure training conditions similar to those of TransMiter, we adapt the transferred soft prompts to the pre-trained strong VLM in an unsupervised training for one epoch using knowledge distillation, with the fine-tuned weak VLM serving as the teacher model.
We set the learning rate to 0.002 and 0.0025 using SGD with a cosine annealing schedule for CoOp~\cite{zhou2022coop} and PromptSRC~\cite{promptsrc}, respectively. 
For all datasets, we use a batch size of 4, except for Imagenet~\cite{imagenet} where we used a batch size of 16.

For the other supervised fine-tuning methods compared and applied in Table 3, such as Maple~\cite{khattakMaPLe}, HPT~\cite{hpt}, and LwEIB~\cite{lweib}, we use ViT-L/14 as the base model and strictly follow the original hyperparameter configurations described in each paper and codebase.

\subsection{Implementation details of TransMiter}
The MLP layer in TransMiter consists of [\texttt{layernorm}-\texttt{linear}-\texttt{activation}-\texttt{linear}], with zero-initialization applied to the last linear layer to preserve the pre-trained knowledge.
We employ orthogonal parameterization techniques~\cite{v2kd} using matrix exponentiation, removing the need for additional orthogonality constraints in the loss function.
For the hyperparameters, we set the dimensionality of the latent space $D$ and logits $M$ to 1024, with the bottleneck dimensionality of the MLP to four times that of the latent space as default.
To train TransMiter, we set the learning rate and weight decay to $1\times 10^{-3}$ and trained for 10 epochs on all datasets, using the AdamW~\cite{adamw} optimizer.
Since TransMiter does not require access to the internal VLM parameters, we extract image and text features before training to reduce computational costs.
For the data augmentation, we applied Gaussian noise to the input logits.
The temperature $\tau_\text{ut-s}$ and $\tau_\text{ft-s}$ were set to 0.01 and 0.005, respectively.
The regularization weight $\beta$ is set to 500 as default.

To further fine-tune TransMiter using labeled data, we train it for 10 additional epochs with a learning rate of $2 \times 10^{-5}$ and a batch size 16, while maintaining all other settings unchanged.
We follow the 16-shot fine-tuning setting, as in the fine-tuning approaches we have adopted, such as CoOp and PromptSRC.

\subsection{Hardware setup}
All the baselines with a training-based approach are conducted on a single 48GB A6000 GPU, as it meets the minimum specifications required for the experiment. 
In contrast, TransMiter demonstrates exceptional efficiency, as all experiments are performed on a single 11GB 2080Ti GPU, except for the computation cost calculation, which is conducted on the A6000 GPU to ensure a fair comparison with the baselines.

\section{Additional experiments}
In this section, we conduct additional experiments including \textbf{(1)} additional baseline and model combination, \textbf{(2)} Transfer across different pre-training schemes, \textbf{(3)} extreme low shot settings, \textbf{(4)} domain generalization, \textbf{(5)} larger model and \textbf{(6)} object detection .
We utilize ViT-B/16 and ViT-L/14 as the source and target models, respectively, with PromptSRC~\cite{promptsrc} as the default fine-tuning strategy unless specified.

\subsection{Additional baseline and model combination}
\begin{table}[ht]
    \centering
    \resizebox{0.25\textwidth}{!}{
    \footnotesize
    \begin{tabular}{c c c}
    \toprule
    \multirow{1}{*}{\textbf{Model}} & \multirow{1}{*}{\textbf{Method}} & Avg. \\
    \midrule
    \midrule
    \rowcolor{cyan!10} 
    \cellcolor{white}{Source} & Fine-tuning  & 73.39 \\
    \midrule
    
    \rowcolor{cyan!10} 
    \cellcolor{white}{\multirow{6}{*}{Target} } & Zero-shot  & 59.60 \\
     & LLMbo &60.21  \\
     & EFT & 71.39 \\
     & Prompt Transfer  & 59.53 \\
    \rowcolor{red!10} 
    \cellcolor{white} & \textbf{TransMiter} & \textbf{74.10} \\
    \bottomrule
\end{tabular}}
    
    \caption{\textbf{Comparison with baselines using ResNet models in base-to-base setting}. 
    The source model is ResNet-50, while the target model is ResNet-101.
    We adopt CoOp~\cite{zhou2022coop} for fine-tuning strategy.
    }
    \label{tab:llmbo_rn}
\end{table}

In \cref{tab:llmbo_rn}, we conduct an additional experiment in which ResNet-50 and ResNet-101 serve as the source and target models, respectively, under base-to-base adaptation transfer setting.
Also, we compare our method with LLMbo~\cite{LLMbb}, a text-based prompt optimization approach that leverages LLMs (\eg ChatGPT~\cite{chatgpt}).
The text prompt obtained in ResNet-50 by LLMbo is used to prompt ResNet-101.
While TransMiter operates in an unsupervised manner, LLMbo requires labeled data for optimization.
TransMiter achieves the highest average accuracy across 11 datasets, outperforming all baselines.

\begin{table}[ht]
\centering
\begin{minipage}{0.46\linewidth}
    \centering
    \resizebox{\linewidth}{!}{
    \begin{tabular}{l c |cc}
        \toprule
        \textbf{Model}& \textbf{Method} & \textbf{1-shot} & \textbf{4-shot} \\
        \midrule
        \midrule
        Source&\cellcolor{cyan!10} Fine-tuning  &\cellcolor{cyan!10} 69.65 &\cellcolor{cyan!10} 77.19 \\
        \midrule
        \multirow{2}{*}{Target} &EFT                 & 73.25 & 78.14 \\
         &\cellcolor{red!10}TransMiter          & \cellcolor{red!10}\textbf{74.78} & \cellcolor{red!10}\textbf{80.12} \\
        \bottomrule
    \end{tabular}}
    \caption{\textbf{Low-shot performance.}}
    \label{tab:low_shot}
\end{minipage}
\hfill
\begin{minipage}{0.50\linewidth}
    \centering
    \vspace{8pt}
    \resizebox{\linewidth}{!}{
    \begin{tabular}{lc|ccc}
        \toprule
        
        \textbf{Model}&\textbf{Method} & \textbf{Base} & \textbf{Novel} & \textbf{HM} \\
        \midrule 
        \midrule
        Source &\cellcolor{cyan!10} Fine-tuning  & \cellcolor{cyan!10}81.00 & \cellcolor{cyan!10}73.87 & \cellcolor{cyan!10}77.27 \\
        \midrule
        \multirow{2}{*}{Target} &EFT                 & 79.95 & 78.92 & 79.43 \\
        
         &\cellcolor{red!10}{TransMiter}          &\cellcolor{red!10} \cellcolor{red!10}\textbf{82.24} &\cellcolor{red!10} \textbf{81.16} & \cellcolor{red!10}\textbf{81.70} \\
        \bottomrule
    \end{tabular}}
    \caption{\textbf{Transfer across different pre-training objectives.}}
    \label{tab:siglip}
\end{minipage}
\vspace{-5mm}
\end{table}

\subsection{Transfer across different pre-taining schemes.}
In \cref{tab:siglip}, we evaluate the transfer of adaptation knowledge across different pre-training schemes, from CLIP (softmax-based NCE loss) to SigLIP~\cite{siglip} (sigmoid loss).
The results show that TransMiter effectively transfers adaptation knowledge across these heterogeneous pre-training objectives.

\subsection{Extreme low shot.}
In addition to the 16-shot base-to-base adaptation setting, we further evaluate extreme few-shot scenarios, where weak VLMs are fine-tuned with only 1 and 4 shots.
As shown in \cref{tab:low_shot}, TransMiter consistently outperforms the baselines even under these low-resource conditions.

\subsection{Domain generalization.}
\begin{table}[ht]
    \centering
    \resizebox{0.42\textwidth}{!}{
    \footnotesize
    \begin{tabular}{c c c c c c c c}
        \toprule
        \multirow{1}{*}{\textbf{Model}}&
        \multirow{1}{*}{\textbf{Method}}
        & -V2 & -S & -A & -R & Avg \\

        \midrule
        \midrule
        \rowcolor{cyan!10}
        
        \cellcolor{white}{Source} & 
        Fine-tuning  & 65.5& 49.3& 48.6& 77.2&60.1 \\
        \midrule
        
        \rowcolor{cyan!10}
        
        \cellcolor{white}{\multirow{5}{*}{Target}} & 
        Zero-shot  & 67.8& 57.8& 68.8& 85.4& 70.0\\
         &
         EFT & 69.6& 57.7& 68.5& 86.3& 70.5\\
         &
         Prompt Transfer  & 66.8& 57.5& 68.3& \textbf{86.5}&69.8 \\

        \rowcolor{red!10}
        \cellcolor{white}&
        \textbf{TransMiter} & \textbf{70.9}& \textbf{59.6}& \textbf{70.1}& \textbf{86.5}&\textbf{71.8} \\



        \bottomrule
    \end{tabular}}
    
    \caption{\textbf{Performance on domain generalization}.
    }
    \label{tab:domain_gen}
\end{table}
In \cref{tab:domain_gen}, we also explore the domain generalization setting by testing the model on 4 out-of-distribution variants of ImageNet, \eg ImageNet-V2 (-V2)~\cite{imgv2}, ImageNet-Sketch~\cite{imagesketch}, ImageNet-A~\cite{imageA}, and ImageNet-R~\cite{imageR}.
For the fine-tuning method, we utilize the 16-shot labeled ImageNet dataset. 
Entire unlabeled ImageNet data is used for the training-based transferable adaptation method.
TransMiter achieves the highest average accuracy among the baselines, demonstrating its effectiveness in handling out-of-distribution scenarios.

\noindent\subsection{Scalability of TransMiter.}
\begin{table}[ht]

    \centering
    \resizebox{0.43\textwidth}{!}{
    \footnotesize
    \begin{tabular}{c c | c c c c}
        \toprule
        \multirow{2}{*}{\textbf{Model}}&
        \multirow{2}{*}{\textbf{Method}}
        & \multicolumn{3}{c}{\textbf{Base-to-Novel}} 
        & \multirow{2}{*}{\textbf{Base-to-Base}} \\
         \cmidrule(r){3-5}
         
           & & Base & Novel & HM & \\
        \midrule
        \midrule
        \rowcolor{cyan!10}
       \cellcolor{white}{ Source}& Fine-tuning& 87.15& 81.59& 84.28&85.17 \\
        \midrule
        \rowcolor{cyan!10}
        \cellcolor{white}{\multirow{3}{*}{Target} } &Zero-shot & 86.95&{87.80} &87.37 &82.12 \\
        
         & EFT & 59.10& 54.35& 56.63& 51.52 \\ 
        \rowcolor{red!10}
       \rowcolor{red!10}
       \cellcolor{white} & \textbf{TransMiter}&\textbf{90.42} & \textbf{87.92}& \textbf{89.15}& \textbf{87.15}\\

        \bottomrule
    \end{tabular}}
    \vspace{-2mm}
    \caption{\textbf{Application on strongest model}. 
    The source model is ViT-L/14, while the target model is ViT-H/14, pre-trained on DFN-5B~\cite{dfn}. 
    }
    \label{tab:large_model}
\end{table}

To show the scalability of TransMiter, we conduct an experiment using ViT-H/14~\cite{dfn} as the target model, one of the strongest open-source VLMs~\cite{openclip} to date, in  \cref{tab:large_model}.
TransMiter with target model surpasses Source Fine-tuning and Target Zero-shot significantly in all adaptation transfer settings.
Interestingly, EFT~\cite{Emulator,proxy_tuning} exhibits a significant drop in performance, indicating that it suffers from a scalability issue.
The above results underscore the applicability of TransMiter, which can effectively enhance huge models.

\noindent \subsection{Application on object detection.}

\begin{table}[ht]
    \centering
    \footnotesize
    \setlength{\tabcolsep}{3pt}
    
    \centering
    \begin{tabular}{l c |c  c c  }
    \toprule
    
    
    
    {\textbf{Model}} &{\textbf{Method}} &  {$\textbf{30 shot}$} & $\textbf{20 shot}$ & $\textbf{10 shot}$ \\
    
    \midrule
    \midrule
    
    \multirow{2}{*}{Source}
    &Zero-shot & 42.7 & 42.7 & 42.7 \\
       &  \cellcolor{cyan!10}{Fine-tuning}& \cellcolor{cyan!10}{45.7}& \cellcolor{cyan!10}{44.9}&\cellcolor{cyan!10}{44.5} \\
    \midrule
    \rowcolor{cyan!10}
    \multirow{2}{*}{\cellcolor{white} Target} &Zero-shot & 52.3 & 52.3 & 52.3 \\
    
     &\cellcolor{red!10}{TransMiter} & \cellcolor{red!10}\textbf{54.4} & \cellcolor{red!10}\textbf{53.7} & \cellcolor{red!10}\textbf{53.3}  \\

    \bottomrule
\end{tabular}
    \vspace{-2mm}
    \caption{\textbf{Few-shot object detection on COCO.} Performance is evaluated using Average Precision (AP).}
    \label{table:od}
\end{table}
In \cref{table:od}, we further evaluate the generality of TransMiter by extending our experiments beyond visual recognition to object detection.
We use ground-truth bounding boxes to crop individual object regions, which are then fed into the VLMs as input images.
The source model is fine-tuned on 10/20/30-shot COCO datasets, where each shot is sampled with sufficiently large object regions.
The results show that TransMiter with the target model consistently outperforms all baselines, demonstrating its broader applicability to dense prediction.

\section{Additional ablations and analyses.}
In this section, we conduct detailed ablations and analyses of TransMiter, including \textbf{(1)} ablation on anchor quantity \textbf{(2)} auxiliary class sampling strategy, \textbf{(3)} latent dimensionality, \textbf{(4)} basis change, \textbf{(5)} transition matrix, \textbf{(6)} regularization weight, \textbf{(7)} analysis on adaptation knowledge, \textbf{(8)} component-wise transfer time, and \textbf{(9)} comparison with PromptKD. 

\subsection{Effect of anchor quantity.}

\begin{table}[ht]
    \centering
    \setlength{\tabcolsep}{3pt}
    \resizebox{0.98\linewidth}{!}{
    
    \centering
    \footnotesize
    \begin{tabular}{l c | c  | c  | c| c}
    \toprule
    
    \multirow{2}{*}{\textbf{Model}} & \multirow{2}{*}{$M$} & DTD ($\Delta$)& UCF ($\Delta$) & Cars ($\Delta$) & SUN ($\Delta$) \\
    && 47 & 100 & 196 & 397\\
    \midrule
    \midrule
    \multirow{4}{*}{Source} 
    & $N_\text{task}$ & 69.82, 0.00 & 83.84, 0.00 & 79.62, 0.00 & 76.28, 0.00\\
    & 256 & 71.77, \textcolor{blue}{+1.95} & 84.30, \textcolor{blue}{+0.46} & 79.96, \textcolor{blue}{+0.34} & - \\
    & 512 & 72.32, \textcolor{blue}{+2.50} & 84.32, \textcolor{blue}{+0.48} & 79.87, \textcolor{blue}{+0.25} & 76.07, \textcolor{red}{-0.21} \\
    & 1024 & 72.54, \textcolor{blue}{+2.72} & 84.37, \textcolor{blue}{+0.53} & 80.52, \textcolor{blue}{+0.90} & 76.46, \textcolor{blue}{+0.18} \\
    \midrule
    \multirow{4}{*}{Target} 
    & $N_\text{task}$ & 64.18, 0.00 & 81.96, 0.00 & 81.38, 0.00  & 76.52, 0.00  \\
    & 256 & 71.91, \textcolor{blue}{+7.73} & 84.25, \textcolor{blue}{+2.29} & 81.47, \textcolor{blue}{+0.09} & - \\
    & 512 & 73.64, \textcolor{blue}{+9.46} & 85.15, \textcolor{blue}{+3.19} & 83.03, \textcolor{blue}{+1.65} & 76.79, \textcolor{blue}{+0.27} \\
    & 1024 & 74.21, \textcolor{blue}{+10.03} & 85.52, \textcolor{blue}{+3.56} & 83.11, \textcolor{blue}{+1.73} & 77.14, \textcolor{blue}{+0.62} \\
    \bottomrule
\end{tabular}}
    \vspace{-2mm}
    \caption{\textbf{Analysis on the number of anchors $M$.} 
    $\Delta$ indicates the performance gap for each $M$ compared to the case where $M$ is set to $N_\text{task}$.
    The number under each dataset name represents the number of task classes for that dataset.
        }
    \label{table:anchor_quant}
\end{table}
In \cref{table:anchor_quant}, we examine the effects of anchor quantity $M$ on both source and target models within a base-to-base setting. 
We observe that increasing $M$ enhances the performance of both the source and target models. 
Notably, the performance gain from additional anchors for the target model is relatively higher across all reported datasets than for the source model.
Further, the performance increases with more anchors.
These results underscore two main implications: (1) incorporating auxiliary classes enhances representational capacity, leading to more effective training of TransMiter and (2) they play a crucial role in achieving high transferability.


\begin{figure}[ht]
\begin{minipage}[t]{0.45\linewidth}
    \centering
    \footnotesize
    \resizebox{\linewidth}{!}{
    \begin{tabular}{l c c c }
        \toprule
        Sampling & Base & Novel & HM \\
        \midrule
        \midrule
        Random & 85.27 & 81.07 & 83.12 \\
        FPS    & \textbf{85.39} & \textbf{81.14} & \textbf{83.21} \\
        Top-k  & 84.98 & 80.98 & 82.93 \\
        \bottomrule
    \end{tabular}}
    \vspace{-1mm}
    \captionof{table}{\textbf{Auxiliary class sampling strategy}.}
    \label{tab:anchor_selection}
    
\end{minipage}
\hfill
\begin{minipage}[t]{0.48\linewidth}

\centering
    \footnotesize
    \resizebox{\linewidth}{!}{
    \begin{tabular}{l c c c }
        \toprule
        $D$ & 512 & 1024 & 2048 \\
        \midrule
        \midrule
        HM & 82.97 & \textbf{83.12} & 82.69 \\
        
        \bottomrule
    \end{tabular}}
    \vspace{-1mm}
    \captionof{table}{\textbf{Latent dimensionality} $D$.}
    \label{tab:latent_dim}
    
\end{minipage}
\vspace{-3mm}
\end{figure}

\subsection{Ablation on auxiliary class sampling strategy.}
Table~\ref{tab:anchor_selection} shows the performance of different auxiliary class sampling strategies. 
FPS (Farthest Point Sampling) selects auxiliary classes that are not only far from the task classes but also mutually distant in the text embedding space, encouraging diversity across the sampled set.
Top-k, on the other hand, selects auxiliary classes most similar to the task classes based on text embedding similarity.
As the choice of sampling strategy introduces minimal variance, consistent with the results in RelRep~\cite{relrep}, we use random sampling for all the experiments for simplicity.

\subsection{Sensitivity analysis on latent dimensionality.}
Table~\ref{tab:latent_dim} presents the impact of varying the latent dimensionality $D$.
Overall, $D=$1024 appears to be the best choice in terms of performance.

\subsection{Ablation on basis change.}

    

\begin{table}[ht]
    \centering
    \setlength{\tabcolsep}{9pt} 
    \footnotesize
    \begin{tabular}{c c| c c c }
        \toprule
         \multicolumn{2}{c|}{\textbf{Method}} & \multicolumn{3}{c}{\textbf{Base-to-Novel}} \\
        \multirow{2}{*}{Feature}
        & \multirow{2}{*}{\shortstack{Transposed \\ mapping}}
        &\multirow{2}{*}{Base} & \multirow{2}{*}{Novel} & \multirow{2}{*}{HM} \\
        &  &  &  &  \\
         \midrule
         \midrule
         \rowcolor{cyan!10}
          \multicolumn{2}{c|}{Zero-shot} & 76.71 &80.85 & 78.73\\
         \midrule
          $h$&      & 85.48& 79.25& 82.25\\
          $\hat{h}$&      & \textbf{85.66}&79.57 &82.50 \\
          \rowcolor{red!10}
          $h$  &\checkmark     & 85.27&\textbf{81.07} &\textbf{83.12} \\

        \bottomrule
    \end{tabular}
    \caption{\textbf{Analysis on basis change}. 
    “Feature” indicates the feature used for obtaining the mapping matrix $\hat{W}$. 
    “Transposed mapping” refers to multiplying the output features $\hat{h}$ by the transposed mapping matrix $\hat{W}^\intercal$.
    }
    \label{table:mapping}
    \vspace{-10pt}
\end{table}

In \cref{table:mapping}, we analyze the basis change when transferring TransMiter from a weaker to a stronger model under the base-to-novel setting.
When using input features $h$ for the alignment, basis change without transposed mapping achieves a significant accuracy gain of 8.77\% and 3.46\% in Base and HM, respectively, compared to the zero-shot performance of the target model.
However, Novel accuracy drops by 1.6\%.
This trend similarly appears even when feature alignment is performed using the output feature $\hat{h}$.
By applying transposed mapping, TransMiter maintains Novel accuracy at the level of target model's zero-shot performance and exceeds it in Base by a margin of 8.56\%, achieving the highest accuracy in HM of 83.01\%.
This demonstrates that maintaining the inverse relationship between the projection and reconstruction matrices plays a crucial role in preserving the strong model’s generalization ability.

\subsection{Ablation on a transition matrix.}
\begin{table}[ht]
    \centering
    \setlength{\tabcolsep}{2pt} 
    \resizebox{0.48\textwidth}{!}{
    \footnotesize
    \begin{tabular}{c c c| c c c c}
        \toprule
        \multirow{2}{*}{\textbf{Model}} 
        & \multirow{2}{*}{$W_\text{in}=W_\text{out}^\intercal$}
        & \multirow{2}{*}{$W^\intercal W=I$}
        & \multicolumn{3}{c}{\textbf{Base-to-Novel}} 
        & \multirow{2}{*}{\textbf{Base-to-Base}} \\
         \cmidrule(r){4-6}
         
         &  & & Base & Novel & HM & \\
        \midrule
        \midrule
        \multirow{4}{*}{Source} 
        &  &  &     83.92 &75.82  &79.66  & 81.33 \\
        & \checkmark &  &83.86&75.86    &79.66  &81.26 \\
        &  & \checkmark & 83.59   & 75.68 & 79.44 &81.16  \\
        & \checkmark & \checkmark &  83.83 &76.21  &79.84 &81.39  \\
        
        \midrule
        \midrule
        
        \multirow{4}{*}{Target} 
        &  & &  83.29 &79.98  &81.60 & 79.77\\
        & \checkmark &   &83.35  & 80.12 &81.70 & 80.20\\
        &  & \checkmark & 81.12   &77.91  &79.48 &76.91  \\
        \rowcolor{red!10}
        \cellcolor{white}& \checkmark & \checkmark & \textbf{85.27} & \textbf{81.07} & \textbf{83.12} & \textbf{82.51}\\


        \bottomrule
    \end{tabular}}
    
    \caption{\textbf{Ablation on a transition matrix} 
    }
    \label{tab:abl_ortho}
    \vspace{-2mm}
\end{table}

\cref{tab:abl_ortho} presents an ablation study on the configuration of the transition matrix. 
We evaluate two conditions: sharing parameters between projection and reconstruction matrix ($W_\text{in}=W_\text{out}^\intercal$) and enforcing orthogonality on each matrix ($W^\intercal W=I$).
When training TransMiter with the source model, parameter sharing with an orthogonal constraint yields the highest performance in both the base-to-novel and base-to-base settings. 
However, the improvement over other configurations is marginal, with only a 0.40\% increase in HM for base-to-novel and a 0.23\% increase for the base-to-base setting compared to the lowest-performing one.

However, when transferring TransMiter to the target model, performance varies significantly depending on the configuration of the transition matrix.
The highest results are achieved when both parameter sharing and orthogonality are applied, yielding an HM of 83.12\% and a base-to-base accuracy of 82.51\%.
Without parameter sharing, the HM for base-to-novel drops to 79.48\%, with a base-to-base performance of 76.91\%. 
Removing only the orthogonality constraint improves performance slightly, yet still falls short of using shared matrices with an orthogonal constraint.
These findings underscore the importance of applying moderate regularization to the transition matrix while ensuring the invertibility between the projection and reconstruction matrices.


\begin{figure}[h]
\begin{minipage}[t]{0.48\linewidth}
        \centering
    \includegraphics[width=\linewidth]{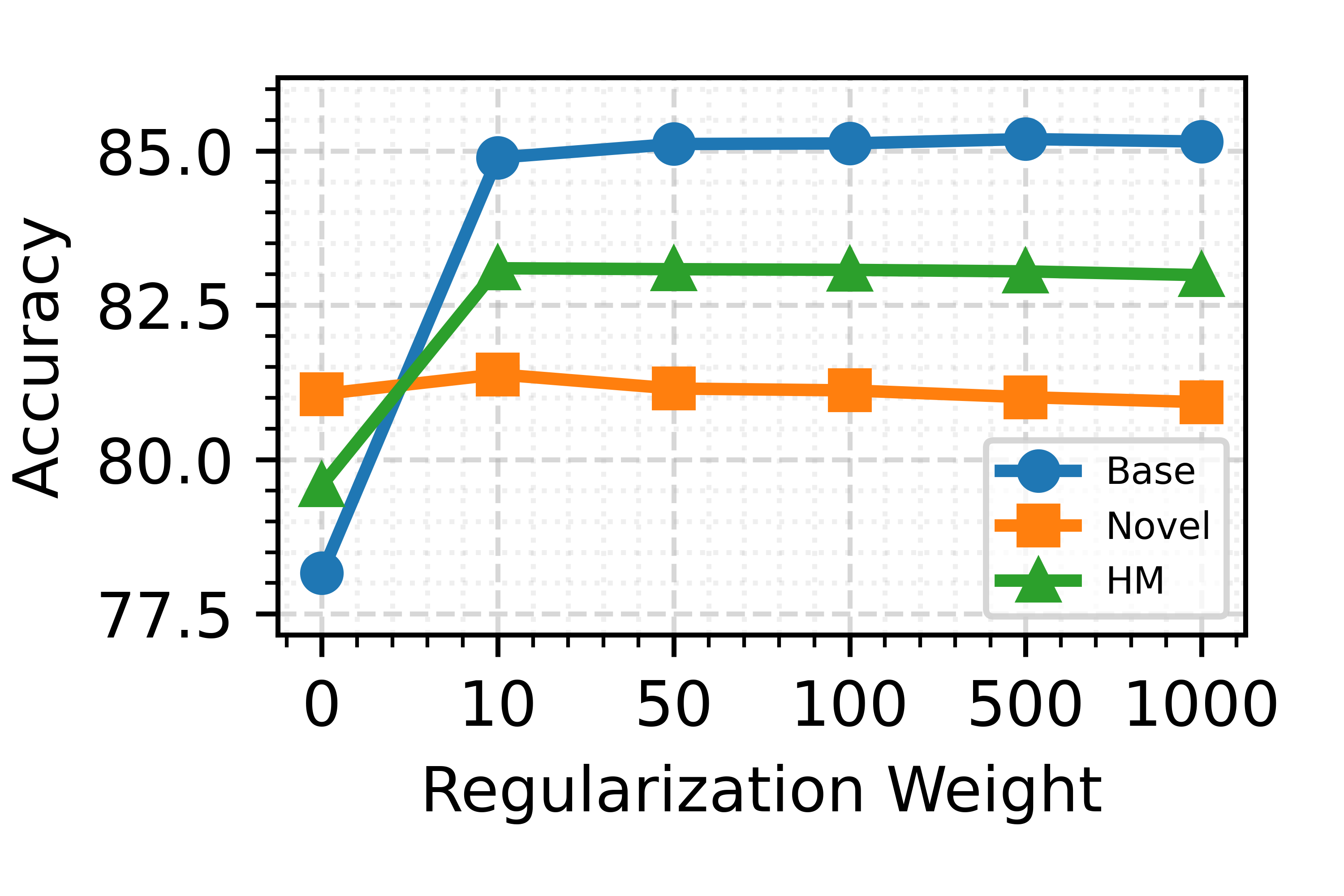}
    \vspace{-8mm}
    \captionof{figure}{\textbf{Regularization weight in basis change.}}
    \label{fig:reg_coef}

\end{minipage}
\hfill
\vspace{5mm}
\begin{minipage}[t]{0.48\linewidth}
    \centering
    \includegraphics[width=\linewidth]{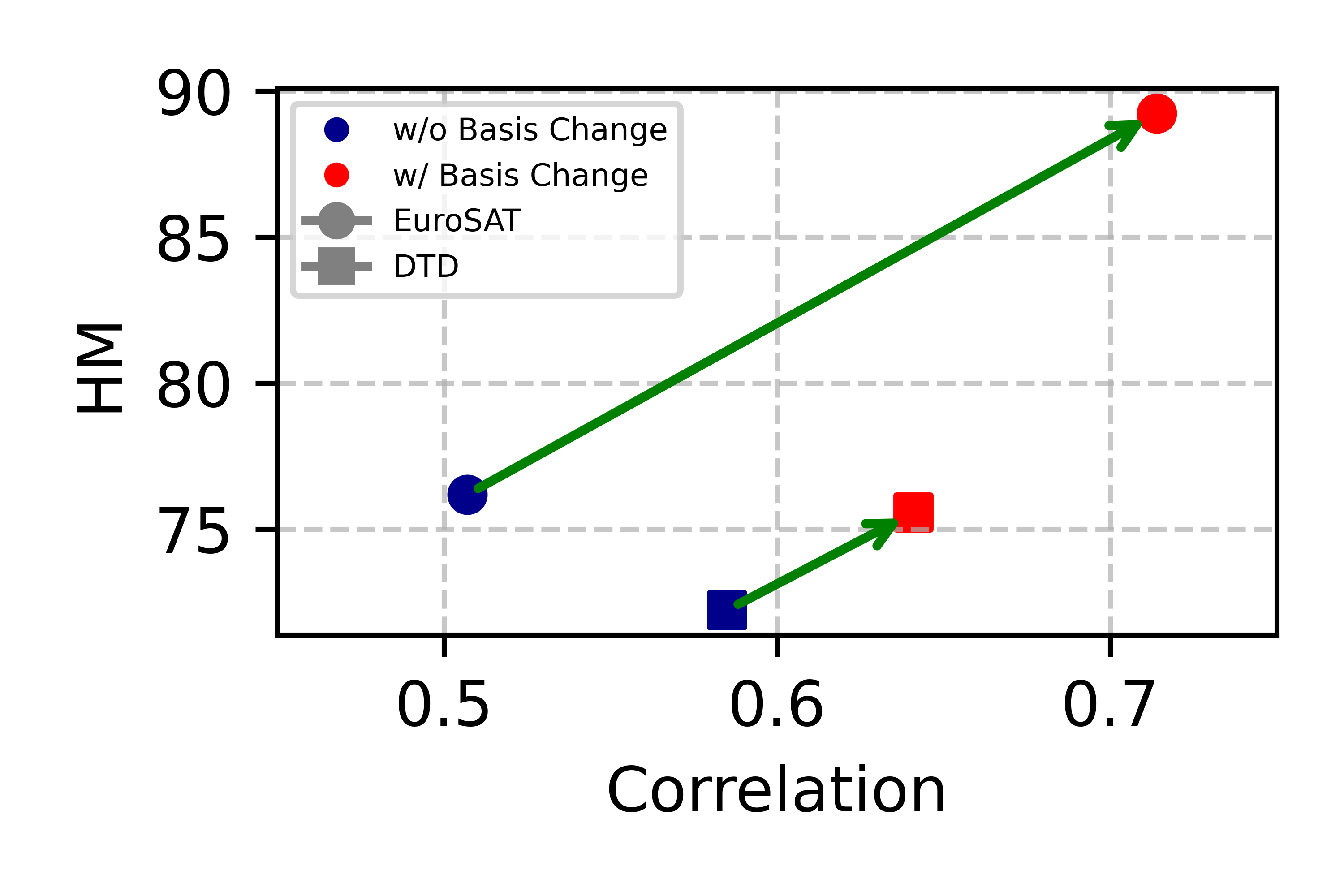}
    \vspace{-8mm}
    \captionof{figure}{\textbf{Adaptation knowledge analysis.}}
    \label{fig:correlation}
    
\end{minipage}
\vspace{-15pt}
    \end{figure}

\subsection{Sensitivity analysis on Regularization weight.}
We conduct a sensitivity analysis on the regularization weight $\beta$ used in basis change.
As shown in \cref{fig:reg_coef}, applying regularization plays a key role in effective adaptation knowledge transfer, and yields stable performance when $\beta>0$.

\subsection{Analysis on adaptation knowledge.}
Our framework transfers the adaptation knowledge, a learning trajectory obtained during fine-tuning a weak model, to a stronger model without using labels. 
Specifically, we extract the logit difference between zero-shot and fine-tuned weak models ($\Delta_s$) (Eq.2, main) and transfer it to the stronger model with backpropagation-free alignment (Eq.3, main).
To verify the effect of our alignment (Basis change), we measure the correlation between the source model's logit shift $\Delta_s={z}_\text{ft-s}-z_\text{pt-s}$ and the target model’s logit shift $\hat{\Delta}_t=\hat{z}_\text{t}-z_\text{pt-t}$ after transfer in \cref{fig:correlation}. 
We observe that applying a basis change between the source and target models enhances the correlation of their learning trajectories, leading to improved performance.
This result highlights the importance of alignment between models for effective adaptation knowledge transfer, thereby validating the necessity of our approach.

\subsection{Component-wise transfer time analysis}

    

\begin{table}[ht]
    \centering
    \resizebox{1.0\linewidth}{!}{
    \begin{tabular}{lcc}
        \toprule
        \textbf{Method}  & \textbf{Transfer time (s)}  &{HM}\\
        \midrule
        \midrule
        Prompt Transfer& 480 &77.21 \\
        PromptSRC & 321& 84.42\\
        
        \midrule
        \rowcolor{gray!10}
        \textbf{Adatptation Knowledge Extraction} & & \\
        TransMiter (Weak VLM feature extraction) & 8& \\
        TransMiter (Fine-tuned weak VLM feature extraction) & 9& \\
        TransMiter (Adaptation knowledge extraction (Eq. 7)) & 24 & 79.52\\
        \rowcolor{gray!10}
        \textbf{Adatptation Knowledge Transfer} & & \\
        TransMiter (Strong VLM feature extraction) & 22  &  \\
        TransMiter (Closed-form (Eq. 11)) & 2 & 81.84 \\
        TransMiter+ (Supervised fine-tuning) & 28 & \textbf{85.16} \\
        \bottomrule
    \end{tabular}}
    
    \caption{\textbf{Detailed transfer time analysis.} The run time is measured on StanfordCars dataset using ViT-L/14 as the base model for all methods.
    Weak and strong VLMs are ViT-B/16 and ViT-L/14, respectively.}
    \label{tab:detail_eff}
    \vspace{-5mm}
\end{table}

In addition to the overall efficiency analysis in Table 6 of the main paper, we provide a component-wise analysis of TransMiter's transfer time in \cref{tab:detail_eff}.
We utilize PromptSRC for fine-tuning a weak VLM.
As we pre-extract features from zero-shot and fine-tuned weak VLMs to reduce TransMiter's training cost, requiring only 17 seconds. 
The knowledge extraction step to capture the knowledge gap adds 24 seconds, totaling just 41 seconds. 
Note that this process is performed only once, and can be reused for any stronger VLM whenever it is introduced.
During adaptation knowledge transfer, most of the runtime is spent on extracting image features from the target model (ViT-L/14) for the basis change.
The closed-form alignment is executed with negligible computational cost.
When incorporating a small amount of labeled data, the supervised fine-tuning completes in only 28 seconds and surpasses its supervised counterpart (PromptSRC), highlighting the practicality of our approach.

\subsection{Comparison with PromptKD.}
\begin{table}[ht]
    \centering
    \resizebox{0.44\textwidth}{!}{
    \footnotesize
    \begin{tabular}{c c c c c}
    \toprule
    \multirow{2}{*}{\textbf{Method}}
    & \multicolumn{3}{c}{\textbf{Base-to-novel}}
    & \multirow{2}{*}{\textbf{Base-to-base}}\\
    \cmidrule(r){2-4}
     & Base & Novel & HM&   \\
    \midrule
    \midrule
    PromptKD & 83.34 & 74.61 & 78.43  &  78.75 \\
    \rowcolor{red!10} 
    \textbf{TransMiter} & \textbf{85.27}& \textbf{81.07}& \textbf{83.01}&\textbf{82.51}   \\
    \bottomrule
\end{tabular}}
    \vspace{-0.2cm}
    \caption{\textbf{Comparison with PromptKD under the weak-to-strong generalization setting}. 
    The source (teacher) and target (student) models are ViT-B/16  and ViT-L/14, respectively. 
    }
    
    \label{tab:sota}
    \vspace{-0.2cm}
\end{table}

\begin{table}[ht]
    \centering
    \resizebox{0.47\textwidth}{!}{
    \footnotesize
    \begin{tabular}{c c c c c }
        \toprule
        \multirow{2}{*}{\textbf{Method}}
        & \multicolumn{2}{c}{\textbf{Transfer}}
        & \multicolumn{1}{c}{\textbf{Inference}}
        & \multirow{2}{*}{\textbf{Acc.}}\\
          & Time (min) & Mem. (Mb)  & FPS&  \\

        \midrule
        \midrule
        
        PromptKD & 3363 & 40820 & 258 & 73.10 \\
        \rowcolor{red!10}
        \textbf{TransMiter} w/o BC& 0& 0& 285 &\textbf{77.05}\\
        \rowcolor{red!10}
        \textbf{TransMiter} w/ BC& 83 & 3784 & 285 &\textbf{77.11}\\



        \bottomrule
    \end{tabular}}
    \caption{\textbf{Efficiency analysis on the ImageNet.} 
    BC refers to Basis Change.
    }
    \vspace{-0.3cm}
    \label{tab:eff_promptkd}
\end{table}

\begin{table}[!ht]
    \centering
    \resizebox{0.47\textwidth}{!}{
    \footnotesize
    \begin{tabular}{ c| c c c |c}
    \toprule
    \textbf{Teacher Method} & Base & Novel & HM & Training time (min) \\
    \midrule
    \midrule
    PromptSRC & \textbf{86.96} & 80.73 & 83.73 &3363\\
    \cellcolor{red!10}\textbf{TransMiter+} & \cellcolor{red!10}{86.53}& \cellcolor{red!10}\textbf{82.70}
    &\cellcolor{red!10} \textbf{84.58}  &\cellcolor{red!10} 2450 \\
    \bottomrule
\end{tabular}}
    \vspace{-0.2cm}
    \caption{\textbf{Application on PromptKD}.
    The teacher and student models are ViT-L/14 and ViT-B/16, respectively.
    }
    
    \label{tab:application_on_promptkd}
    \vspace{-0.2cm}
\end{table}
We also compared TransMiter with the other state-of-the-art unsupervised adaptation method, PromptKD~\cite{li2024promptkd} in \cref{tab:sota} and \cref{tab:eff_promptkd}.
PromptKD is similar to Prompt Transfer in that it transfers knowledge from a fine-tuned teacher model to a student model using naive knowledge distillation~\cite{naive_kd}. 
Additionally, it involves transferring specific parameters of the teacher model, such as the text encoder, which were obtained during the fine-tuning.
We strictly adhere to the implementation details provided in PromptKD, with the only difference being that we use a weaker model (\eg ViT-B/16) as the teacher and a stronger model (\eg ViT-L/14) as the student.

As shown in \cref{tab:sota}, TransMiter significantly outperforms PromptKD across all metrics and adaptation transfer settings.
Notably, PromptKD incurs significant transfer time and memory costs due to the need for backpropagation through internal VLM parameters, whereas TransMiter requires negligible or even zero transfer costs.
Additionally, TransMiter enables efficient inference, consisting of only a few stacked layers, whereas deep prompting~\cite{khattakMaPLe} slows inference in PromptKD.
These results further emphasize that, within the ``knowledge extraction and transfer'' framework, TransMiter facilitates knowledge transfer from weaker to stronger models more effectively and efficiently than PromptKD.

In addition, TransMiter, when trained with few-labeled data (TransMiter+), can be effectively applied to strong-to-weak generalization. 
To validate this, we replaced the teacher model in PromptKD’s distillation framework with TransMiter+. 
As shown in \cref{tab:application_on_promptkd}, the teacher trained with TransMiter+ not only improves performance but also reduces training time due to its faster inference speed. 
These results demonstrate that TransMiter is effective not only for weak-to-strong generalization but also for strong-to-weak scenarios.
\section{Derivation of Eq. (10)}
\begin{equation}
    \begin{split}
       \hat{W}&=\argmin_{W\in\mathcal{O}}||H_\text{t} W-H_\text{s}||_2^2+\beta||W-I||_2^2\\
       &=\argmin_{W\in\mathcal{O}} \text{Tr}\left((H_\text{t}W-H_\text{s})^\intercal (H_\text{t}W-H_\text{s})\right) \\&\quad + \beta\text{Tr}\left((W-I)^\intercal (W-I)\right)\\
       &=\argmin_{W\in\mathcal{O}} \text{Tr}\left(W^\intercal H_\text{t}^\intercal H_\text{t} W\right)-2\text{Tr}(W^\intercal H_\text{t}^\intercal H_\text{s} )\\
       &\quad +\text{Tr}(H_\text{s}^\intercal H_\text{s}) +\beta\text{Tr}(W^\intercal W) - 2\beta\text{Tr}(W^\intercal) \\
       &=\argmax_{W\in\mathcal{O}} \text{Tr}(W^\intercal H_\text{t}^\intercal H_\text{s})+\beta \text{Tr}(W^\intercal)\\
       &=\argmax_{W\in\mathcal{O}} \text{Tr}(W^\intercal(H_\text{t}^\intercal H_\text{s}+\beta I))\\
    \end{split}
    \label{eq:proof1}
\end{equation}

Let $[U,\Sigma,V]=\text{SVD}(H_\text{t}^\intercal H_\text{s}+\beta I)$. Then
\begin{equation}
    \begin{split}
       \text{Tr}(W^\intercal(H_\text{t}^\intercal H_\text{s}+\beta I))
       &= \text{Tr}(W^\intercal U\Sigma V^\intercal )\\
       &= \text{Tr}(V^\intercal W^\intercal U \Sigma  )\\
       &=\text{Tr}(S \Sigma ) \quad \text{(where $S=V^\intercal W^\intercal U$)}\\
       &=\sum_{i}S_{i,i} \Sigma_{i,i}\\
       &\leq\sum_{i}\Sigma_{i,i}.\\
    \end{split}
    \label{eq:proof2}
\end{equation}
Since $S$ is an orthonormal matrix, the last inequality holds and the objective $\text{Tr}(W^\intercal(H_\text{t}^\intercal H_\text{s}+\beta I))$ is maximized when $S=V^\intercal W^\intercal U=I$.
Thus,  $W = UV^\intercal$.

\clearpage

\end{document}